\documentclass[sn-mathphys]{sn-jnl}% Math and Physical Sciences Reference Style
\jyear{2022}%

\theoremstyle{thmstyleone}%
\usepackage[font={footnotesize}]{caption}
\usepackage[para,online,flushleft]{threeparttable}
\usepackage{amsmath,amssymb,amsfonts}
\usepackage{graphicx}
\usepackage{textcomp}
\usepackage{xcolor}
\usepackage{hyperref}
\usepackage{url}
\usepackage{wrapfig}
\usepackage{algpseudocode}

\usepackage{booktabs}

\usepackage{adjustbox}
\usepackage{subfig}
\usepackage{multirow}
\usepackage{marvosym}
\usepackage{dblfloatfix}

\theoremstyle{thmstyletwo}%

\theoremstyle{thmstylethree}%

\begin{document}

\title[Understanding CNN Fragility]{\textbf{Understanding CNN Fragility When Learning With Imbalanced Data}}


\author*[1]{\fnm{Damien} \sur{Dablain}}\email{ddablain@nd.edu}

\author[2]{\fnm{Kristen N.} \sur{Jacobson}}\email{kristen.jacobson@nrl.navy.mil}

\author[3]{\fnm{Colin} \sur{Bellinger}}\email{colin.bellinger@nrc-cnrc.gc.ca}

\author[2]{\fnm{Mark} \sur{Roberts}}\email{mark.roberts@nrl.navy.mil}

\author[1]{\fnm{Nitesh} \sur{Chawla}}\email{nchawla@nd.edu}

\affil*[1]{\orgdiv{Dept. Computer Science and Engineering}, \orgname{University of Notre Dame}, \orgaddress{\street{10 Main}, \city{South Bend}, \postcode{46556}, \state{IN}, \country{USA}}}

\affil[2]{\orgdiv{AI Center}, \orgname{U.S. Naval Research Laboratory}, \orgaddress{\street{4555 Overlook Ave.}, \city{Washington}, \postcode{20375}, \state{DC}, \country{USA}}}

\affil[3]{\orgdiv{NRC}, \orgname{National Research Council of Canada}, \orgaddress{\street{1200 Montreal Rd.}, \city{Ottawa}, \postcode{K1A 0R6}, \state{Ontario}, \country{CA}}}


\abstract{Convolutional neural networks (CNNs) have achieved impressive results on imbalanced image data, but they still have difficulty generalizing to minority classes and their decisions are difficult to interpret.  These problems are related because the method by which CNNs generalize to minority classes, which requires improvement, is wrapped in a black-box. To demystify CNN decisions on imbalanced data, we focus on their latent features.  Although CNNs embed the pattern knowledge learned from a training set in model parameters, the \textit{effect} of this knowledge is contained in \textit{feature and classification embeddings} (\textit{FE} and \textit{CE}).  These embeddings can be extracted from a trained model and their global, class properties (e.g., frequency, magnitude and identity) can be analyzed.  We find that important information regarding the ability of a neural network to generalize to minority classes resides in the \textit{class top-K CE and FE}. We show that a CNN learns a limited number of \textit{class top-K CE} per category, and that their number and magnitudes vary based on whether the same class is balanced or imbalanced. This calls into question whether a CNN has learned intrinsic class features, or merely frequently occurring ones that happen to exist in the sampled class distribution. We also hypothesize that latent class \textit{diversity} is as important as the number of class examples, which has important implications for re-sampling and cost-sensitive methods. These methods generally focus on rebalancing model weights, class numbers and margins; instead of diversifying class latent features through augmentation. We also demonstrate that a CNN has difficulty generalizing to test data if the magnitude of its top-K latent features do not match the training set.  We use three popular image datasets and two  cost-sensitive algorithms commonly employed in imbalanced learning for our experiments.}




\maketitle

\section{Introduction}\label{sec1}

CNNs are increasingly being applied to imbalanced visual data in high-stakes fields such as medicine, business and law \cite{johnson2019survey}. Yet, they have difficulty generalizing to classes with few examples.  Imbalanced data helps focus the spotlight on generalization because it provides a contrast between majority classes, with their rich data profile, and emaciated minority classes, with few examples that typically exhibit a more narrow range of variation.  

The ability of a neural network to generalize on minority classes is critical to its overall performance.  For example, in medicine, physicians may be more interested in the accurate recognition of minority instances, such as cancerous lung tissue.  Improving model generalization on minority classes is challenging because neural networks are opaque \cite{gunning2019darpa}.  The black-box nature of CNNs makes it difficult to study the very problem that we are interested in: why does a CNN struggle to generalize on minority classes?  To answer this question, we first have to unravel its decision process so that the properties of the features that cause it to misclassify minority examples can be identified.  Although there are many available explainable Artificial Intelligence (XAI) methods that examine neural network feature relevance, there is a paucity of research that combines and analyzes data imbalance, generalization and CNN opacity in a single, unified study.  XAI feature relevance methods such as LIME \cite{ribeiro2016should} or Shapley values \cite{sundararajan2020many,shapley1997value} generally focus on instance, instead of class, features.  Similarly, pixel attribution methods, such as saliency maps \cite{simonyan2013deep,sundararajan2017axiomatic}, network deconvolution \cite{zeiler2014visualizing}, and activation maps \cite{selvaraju2017grad} focus on attributing CNN predictions on specific images to input pixels, instead of interpreting network decisions rendered on an entire class.

%\noindent \textbf{Summary.} 
\begin{figure*}[t!]
%\small
%\scriptsize
%\centering
\footnotesize
\vspace{-1.2cm}
  \includegraphics[width=0.9\textwidth]{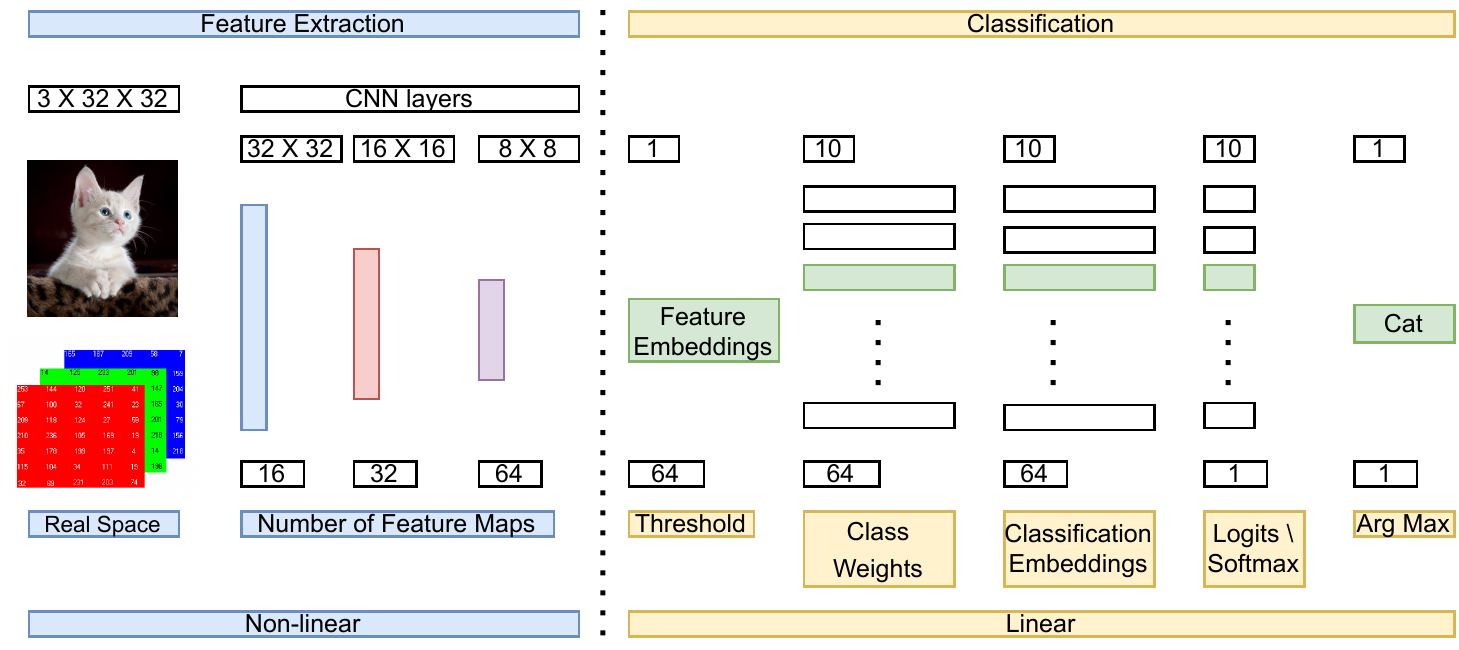}
  
  %\subfloat[Balanced: CIFAR10]{\includegraphics[width=0.9\textwidth]{cnn3.pdf}\label{fig:cnn}}
  \caption{Illustration of feature embeddings (FE) and classification embeddings (CE), using the Resnet 32 architecture. The CNN's extraction layers produce feature maps based on the interaction of convolutional layers and non-linear activations with input pixels.  After thresholding, FE represent a low-dimensional response to the input.  Based on the classification layer's final prediction, we trace the final output (a label) to logits, classification embeddings and feature embeddings that triggered the response.  By comparing the CE of the predicted class to the next largest class logit, we determine the number of relevant FE and CE required to make a prediction (top-K).}
  \label{fig:cnn}
  \vspace{-0.1cm}
\end{figure*}

In this work, we strive to better understand the process by which CNNs reach their decisions on imbalanced image data. To search for an answer to this problem and to make it tractable, we examine the latent representation that CNN's extract from input data and use when making a class prediction. After thresholding, this embedding represents a vector of low dimensional features that a linear classifier uses to predict a label.  We investigate the properties of these latent features (i.e., their magnitude, identity, and frequency), their relationship to class weights, and draw general hypotheses about how CNN's generalize with respect to majority and minority classes. To enable our research, we break a CNN into two separate networks, extraction and classification layers, so that we can concentrate on the latent features that serve as input to the recognition process.  We refer to the internal representation learned by the extraction layers, after thresholding, as \textit{feature embeddings (FE)} and the output of the classification layer, before summation and Softmax, as \textit{classification embeddings (CE)}.  See Figure ~\ref{fig:cnn} for an illustration.

\subsection{Main Contributions}\label{sec_contrib}

We make the following research contributions by taking steps to explain the decision process of CNN's when operating on imbalanced data:
\begin{itemize}
    \item \textbf{CNN minority class latent features are less diverse.}  We measure class feature diversity based on the number, and density, of feature and classification embeddings.  We use mean as a measure of density and show that a CNN's internal representation of minority class features is less diverse than the majority.
    \item \textbf{CNN minority class prediction rests on fewer, higher valued relevant features due to low diversity.}  A CNN's minority class prediction rests on only a handful of features in embedding space (the top-K features), which is of lower size than the embedding dimension, but generally of higher mean magnitude than relevant majority latent features.  There are fewer relevant features for minority classes than majority ones.  We hypothesize that the decision manifold is narrower for minority classes because there is less diversity of examples.  The majority class distribution is more diverse, hence it requires a larger decision manifold (number of relevant features) in latent space to reach a decision (represent a class).  
    
    \item \textbf{Higher response, lower number of minority class features leads to poor generalization.}  Although a CNN classifier relies on fewer relevant features to distinguish minority examples, it compensates by increasing the magnitude of the top-K minority features. This finding is interesting in light of previous work which found that majority classes dominate CNN model gradients \cite{anand1993improved}. We conjecture that this phenomena occurs because, due to fewer examples and less diverse features, the system's response to top-K minority features is elevated to ensure proper classification. This may partially explain why CNN's have difficulty generalizing to minority class examples. The system is conditioned to engineer a high response to a limited number of minority features, and when those high response features are not present in the test set (due to lower minority class \textit{FE} magnitudes spread over more \textit{FE}), the classifier mistakes the minority example for an adversary (majority) class with lower, and more varied, overall response to the input. In contrast, the classifier has been conditioned to expect a wider range of majority class features and hence, each individual feature has a lower magnitude and the sum of more, lower magnitude features allows the classifier to make the correct majority class prediction.
    
    \item \textbf{Generalization capacity.}  A CNN has difficulty generalizing from the training to test set if the range of its latent feature magnitudes differ.  We demonstrate that a CNN is able to generalize from the training to the test set if there is a close match in the range of its top-K \textit{FE} features. 
\end{itemize}

\section{Background \& Related Work}
\label{sec:back_rw}
\subsection{Background}
We assume that CNNs perform visual recognition in a two-step process. First, low dimensional embeddings are extracted. Second, based on these low dimensional features, object classification occurs (a decision is rendered).  This assumption forms the basis of our experiments, where we separate a CNN into two basic layer groups, extraction and classification, so that we can better understand the CNN decision process (classification).

There is some support for this approach. The manifold hypothesis holds that high-dimensional data can be represented on a less complicated, lower dimensional manifold \cite{cayton2005algorithms}. In the computer vision field, it has similarly been hypothesized that high-dimensional image data can be expressed in a more compact form, based on latent features \cite{brahma2015deep,bellinger2018manifold}. 

Many modern CNN architectures, which use a non-linear rectified linear unit (ReLU) activation function, can be viewed as approximating high dimensional image data in a lower dimensional embedding space (with extraction layers) \cite{li2022robust,chui2018deep}. If complex, high dimensional input can be reduced to a low dimensional latent space, then linear models can be more readily applied to reach a decision (in the classification layer). Stated differently, CNNs use non-linearity to find low dimensional features (with extraction layers) that can be linearly separated (by classification layers).

Although CNNs have achieved impressive results, they face key challenges, especially with regard to their ability to generalize on imbalanced data. First, classifiers tend to obtain their highest accuracy when the density of positive and negative examples along a class decision boundary are approximately the same \cite{smote-comparison,huang2019deep}. However, when there is a wide divergence in the number and diversity of class examples, such as with majority and minority classes, the decision boundary can become blurred. 

Second, the decision process of a CNN is difficult to understand, which further compounds the first problem because the decision boundary is wrapped in a black-box \cite{karimi2019characterizing,mickisch2020understanding}. The ML sub-fields of imbalanced learning and XAI \textit{independently} address these issues: improving classifier accuracy for minority classes (imbalanced learning) and model interpretability (XAI). There has been a paucity of research that combines these two approaches into a single, unified discussion. 

\subsection{Related work}
\label{sec:rw}
\textbf{XAI.} XAI adopts a variety of approaches to explain model decisions, including: explaining more complicated models by reference to simpler ones, feature relevance, various post-hoc methods, explanation by example, and mapping predictions to inputs \cite{gunning2019darpa,adadi2018peeking,linardatos2020explainable}. Many of these approaches are local in nature because they explain a model decision on a single input, and do not attempt to explain global properties of features or decision processes \cite{achtibat2022towards}. For example, feature relevance techniques, which are related to our approach, such as Shapley values or Local Interpretable Model-Agnostic Explanations (LIME), show the importance of features to a single instance. Shapley values typically involve retraining a model or modifying data on a single instance to understand feature relevance \cite{sundararajan2020many}. LIME requires learning another model locally around a single prediction \cite{ribeiro2016should}.  In addition, Bau et. al propose network dissection, which evaluates the alignment of individual hidden neurons with semantic concepts \cite{bau2017network}.  Kim et. al propose directional derivatives to quantify the degree to which a concept is important \cite{kim2018interpretability}. All of these works focus on individual predictions, instead of class feature properties.  Badola et. al develop the notion of instance top-K features produced by a CNN filter \cite{badola2021identifying}, although they do not apply this to imbalanced data.  

In addition to XAI, feature relevance has been explored in the ML sub-field of adversarial examples (AE) \cite{szegedy2013intriguing}.  Adversarial examples can generally be described as small perturbations of images (in pixel space) in the direction of the sign of the input gradient.  Ilyas et. al demonstrate that CNNs learn highly predictive, yet brittle, patterns that are not comprehensible to humans \cite{ilyas2019adversarial}. In the same context, Wang et. al show that CNNs learn high frequency patterns that are incomprehensible to humans and which contribute to adversarial examples \cite{wang2020high}.

\textbf{Imbalanced learning.} Imbalanced learning is concerned with designing methods that allow classifiers to better generalize from training to test data on minority classes \cite{fernandez2018learning,he2009learning,buda2018systematic}. It uses a variety of approaches: re-sampling minority and majority class data, cost-sensitive methods that assign a greater loss to minority class misclassification, separating a ML system into extraction and classification phases, ensemble, and hybrid approaches \cite{johnson2019survey,krawczyk2016learning,bellinger2020remix}.  

Kang et. al and Zhou et. al develop a novel technique to improve CNN classification with respect to minority classes - bifurcate the model into two separate layer groups: extraction and classification \cite{kang2019decoupling,zhou2020bbn}.  Their approach is used to improve classifier accuracy and not for explanation.  Ye et. al discuss the concept of feature deviation in imbalanced learning on image data, although they do not analyze the properties of this deviation (e.g., feature magnitude, index identity or frequency) \cite{ye2020identifying}. Cao et. al and Kim et. al discuss the impact of imbalanced classes on decision boundaries, classifier weights and class rebalancing \cite{cao2019learning,kim2020adjusting}, although they do not tie this analysis into latent features. 

\section{Methodology}
\label{sec:meth}

\subsection{Nomenclature}
\label{sec:nomen}
The following nomenclature is used to describe our experimental setup, results and conclusions.

A dataset, $D=\{X,Y\}$ is comprised of instances, $X$, and labels, $Y$. An instance, $d=\{x,y\} \in \{D\}$, consists of an image, where $x \in \mathbb{R}^{cXhXw}$, such that $c$, $h$, and $w$ are image channels, height and width, respectively. $D$ can be partitioned into training and test sets ($D=\{Train,Test\}$).

A CNN can be described as a network of weights, $W$, arranged in layers, $L$, that operate on $x$ to produce an output, $y$ (a label).  We partition the layers, $L$, into two principal parts: extraction layers and a classification layer.  (See Figure ~\ref{fig:cnn} for an illustration.) A CNN can then be expressed as:  $f_\theta(\cdot)= f_{W_C}[(f_{W_E})_{Th}]$, where $f_{W_E}(\cdot)$ are the extraction layers, $Th$ performs thresholding, $f_{W_C}(\cdot)$ is the classification layer, $W_E$ are extraction layer weights, and $W_C$ are classification layer weights.  Feature embeddings (FE) are the output of the extraction layers after thresholding has been applied, or $\mathit{FE} = (f_{W_E})_{Th}$.  Classification embeddings (CE) are the result of the Hadamard product of FE and the transpose of the classification weights, or $CE = FE \cdot W_C.T$.  Logits (LG) represent the row-wise summation of CE, or $LG =  \Sigma(CE)$.  The final prediction (y) is the argmax of the Softmax of the logits, or  $y=argmax(\sigma(LG))$, where $\sigma$ is the Softmax function.  Figure ~\ref{fig:cnn} illustrates this nomenclature for the Resnet-32 architecture.

To distinguish classes in a dataset, $D$, we refer to reference and adversary classes. A reference class is the \textit{ predicted} label and an adversary class is any other class in $C$.  The number of classes in $C$ is referred to as $N_C$, with each class $C=\{c_1,c_2,..c_n\}$. Each individual \textit{FE} and \textit{CE} vector can be described as $\mathit{FE}=\{{fe}_1,{fe}_2,..{fe}_h\}$ and  $\mathit{CE}=\{{ce}_1,{ce}_2,..{ce}_h\}$, respectively. Each $fe$ and $ce$ in a single \textit{FE} or \textit{CE}, respectively, have a fixed index position in a vector.

\subsection{Feature Properties}
Model \textit{FE} can be extracted for all $Train$ and $Test$ instances, along with $W_C$, to facilitate the analysis of class feature properties.  Throughout the text, we discuss and quantify several properties of a CNN's internal embeddings, including their identity, magnitude and frequency. 

The \textit{identity} of a $fe_h$ or $ce_h$ refers to its index position in a vector. The \textit{magnitude} of a $ce_h$ or $fe_h$ refers to its value. The \textit{frequency} of a $fe_h$ or $ce_h$ refers to how often it appears within a class in $Train$ or $Test$.  

These properties allows us to compare the number, size, range, and frequency of \textit{FE} that the model uses to define a class.  By contrasting majority and minority class feature properties, we can better understand the CNN's ability to generalize to the test distribution based on its learned features and class weights.

\subsection{Feature Relevance \& Diversity: Top-K FE}
A CNN classifier's prediction for a single data instance, $x$, is based on whether the logit of the reference class exceeds the next largest logit of an adversary class.  This observation applies to CNN's using cross-entropy loss, or a cost-sensitive variant. The label of a final class prediction represents an index in a vector of size $N_C$.  This index points in a "backward" direction to an index $c$ in \textit{CE}. For a CNN that uses cross-entropy loss,  only the \textit{CE} of the reference class (the prediction) and the next largest \textit{CE}
(largest adversary class) matter because the prediction is the argmax of the summed \textit{CE}.  We refer to the reference class \textit{CE} as $\mathit{CE}_R$ and the \textit{CE} of the largest adversary class as $\mathit{CE}_A$.  The respective logits are $LG_R$ and $LG_A$.

The \textit{top-K ce} of each data instance is then the number of individual $ce$ of the reference class required to exceed the next largest logit, $LG_A$. The ability of a given value of K to predict all instances in $Train$ can be determined experimentally by summing the \textit{top-K ce} for each instance and comparing it to each $LG_A$ and quantifying the percentage of times that the sum exceeds $LG_A$ in $Train$.  We refer to this percentage as the \textit{top-K coverage ratio}. The ratio is bounded by 0 and 1.  For a given K, a high top-K coverage ratio means that only K number of $ce$ are needed to predict a high percentage of instances in a training set.  This same procedure can be applied on a class basis or \textit{class top-K coverage ratio}.

This ratio provides an indication of \textit{feature diversity} when examining classes that are imbalanced. If a minority class can be defined by a small value of K (only a handful of features are present in all class instances), then its \textit{top-K coverage ratio} for the given K should be high (near 1). If a majority class has a low class coverage ratio for the same value of K, then a larger number of features are required to make accurate predictions. 

\textit{Top-K fe} or \textit{top-K ce} are instance based measures.  In other words, they determine the top features per instance; however, the specific identity of the top-K components may vary across all instances in a class.  \textit{Class top-K members} are the group of top-K features that occur most frequently across all instances in a class.  

\subsection{Class Feature Means}
For each class, the mean value of each feature ($\{{fe}_1,{fe}_2,..{fe}_H\}$ and $\{{ce}_1,{ce}_2,..{ce}_H\}$) is instructive because it provides insight into the model's response to a given feature. For example, if the mean value of ${fe}_{35}$ is high for class 0, but low for class 7, then this implies that this feature is more important for purposes of distinguishing class 0.  Because a CNN classifier makes it class selection linearly based on the largest logit, high valued features that compose the logit are important to its decision.  The mean magnitude is also a measure of density. For example, if $ce_1$ has a high mean magnitude for a minority class, but it has a low mean magnitude for a majority class, then it implies that $ce_1$ frequently clusters around a high value for a minority class.

\section{Experimental study set-up}
\label{sec:exp}
  Our goal is to better understand the components of the CNN's decision, $ce$, $W_C$, $fe$, and their properties.  By better understanding their properties, we expose their diversity, how this diversity changes with imbalance, and how diversity may affect generalization.  We first attempt to understand if there is a lower number of features to study (top-K).

\subsection{Data}
\label{sec:data}
To conduct our experiments, we examine three popular image datasets: CIFAR-10 \cite{krizhevsky2009learning}, Street View House Numbers (SVHN) \cite{netzer2011reading}, and CelebA \cite{liu2015faceattributes}. In addition, we compare cross-entropy loss on the CIFAR-10 dataset with two cost-sensitive algorithms on the same dataset - LDAM \cite{cao2019learning} and the focal loss \cite{lin2017focal}. The datasets span three different image data types: objects (CIFAR-10), numbers (SVHN) and facial attributes (CelebA). In addition, by comparing a single dataset (CIFAR-10) trained with different loss functions, we are better able to identify the effects of cost-sensitive algorithms on features.

In our experiments, CIFAR-10, SVHN and CelebA contain 10, 10 and 2 classes, respectively. For CelebA, the two classes are: men and women with black hair. We use a single hair color because the full CelebA dataset disproportionately contains more women with blond hair then men, and we want to avoid a simple feature (hair color) that can easily distinguish classes.

The CIFAR-10 training and test sets are initially balanced. For SVHN, we randomly select 4,600 training and 1,500 test instances because the dataset contains an uneven number of training and test examples by class.  For CelebA, we randomly select 5,000 training and 1,000 test images by class.

For purposes of this study, we introduce exponential imbalance into the training set (maximum imbalance ratio of 100:1), similar to Cao et al. \cite{cao2019learning}, for CIFAR-10 and SVHN. For CelebA, the imbalance ratio is 20:1.   

This approach allows us to train two models with identical architectures and training regimes, but with balanced and imbalanced versions of the same datasets.  We can then more precisely observe the impact of imbalance on class feature and weight selection.  

The use of balanced \textit{test} sets allows us to examine the effect of different training and test distributions for minority classes.  More specifically, in the majority class, we would expect that the training and test feature distributions are likely more uniform, and hence, the model should be able to better generalize from the training to the test set.  In contrast, for minority classes, which have a limited number of training examples, the model will likely struggle to generalize to the test set.  For example, in the case of CIFAR-10, there are 5,000 training and 1,000 test examples for the majority class, but there are only 50 training and 1,000 test examples for the smallest minority class.

\subsection{Model architecture \& training regime}
\label{sec:arch}
For CIFAR-10 and SVHN, a Resnet 32 architecture is used and a Resnet 52 architecture is used for CelebA \cite{he2016deep}.  We follow a popular training regime used in cost-sensitive learning for imbalanced data \cite{cao2019learning}. We train for 200 epochs for CIFAR-10 and SVHN and 50 epochs for CelebA.  All models are trained with PyTorch \cite{paszke2017automatic} on a single RTX 3060 Nvidia GPU.  We assess the performance of our trained models with balanced accuracy (BAC), which treats each class equally, regardless of the number of examples. The epoch with the best performing BAC is then selected.

\subsection{Research Questions (RQ) }
\label{sec:rq1_setup}
We summarize below our research questions:

\begin{itemize}
\setlength{\itemindent}{.6cm}
    \item [RQ1:] Can classifier retraining achieve balanced training performance?  
    \item [RQ2:] How does class imbalance affect minority class generalization? 
    \item [RQ3:] Do CNNs rely on $K<\!\!<\!H$ relevant features when classifying an instance and a class?  Is the number of relevant features affected by class imbalance? 
    \item [RQ4:] Do majority classes exhibit greater diversity of $ce$ features than minority classes? 
    \item [RQ5:] Does the magnitude of classifier weights vary based on class imbalance?
    \item [RQ6:] Is the diversity of class $fe$ affected by imbalance? 
    \item [RQ7:] What do the $fe$ of test set true and false positives tell us about the ability of a CNN to generalize to minority classes?
\end{itemize}

\section{Results}
\label{sec:res}

\subsection{RQ1: can classifier retraining achieve balanced training performance?}
\label{sec:rq1_res}

For our initial experiment, we train two Resnet-32 models: one with balanced data and one with exponentially imbalanced data, using cross-entropy loss (C-ent).  We bifurcate the model trained on \textit{imbalanced} data into extraction and classification layers.  We re-train the imbalanced-model classifier with \textit{FE} from the balanced (full) training set, but that were extracted using \textit{imbalanced  extraction layers}.  This procedure focuses the spotlight on the benefits of classifier re-training.  For a combined CNN extractor and classifier trained on \textit{balanced} CIFAR-10 data, the BAC for all classes is 92.65\%.  For a combined CNN extractor and classifier trained on \textit{imbalanced} data, BAC is 72.56\%.  For a CNN extractor separately trained on imbalanced data and a classifier retrained with balanced data extracted by an imbalanced extractor, BAC is only 78.48\%.  The 78.48\% is approximately the BAC achieved by several recent cost-sensitive and classifier re-balancing methods on an exponentially imbalanced CIFAR-10 dataset \cite{cao2019learning,https://doi.org/10.48550/arxiv.2207.06080}.  As noted in Table \ref{tab: bac}, similar results are produced by the other datasets and cost-sensitive algorithms.  In other words, a classifier retrained with features extracted from an imbalanced extractor cannot recover the accuracy levels of a full CNN (extraction and classification layers) trained on balanced data. This result holds even though the classifier is retrained with features drawn from the \textit{full} dataset (albeit from extraction layers trained on imbalanced data), 

\begin{wraptable}{L}{0.6\textwidth}
%\begin{wrapfigure}{L}{0.5\textwidth}
%\begin{minipage}{0.5\textwidth}
%\vspace{-1cm}
%\begin{table}[h!] %*
\begin{threeparttable}
\vspace{-0.5cm}
%\small
%\scriptsize
\footnotesize
\caption{Re-trained Classifier BAC}
\label{tab: bac}
\centering
\begin{tabular}{ p{2.2cm}p{.9cm}
p{.9cm}p{1.3cm}}
\toprule

\textbf{Description} & \textbf{Bal. Train} & \textbf{Imb. Train} & \textbf{Classifier Re-Train}\\

\midrule

C-ent CIFAR-10 & 92.65 & 72.56 & 78.60 \\
Focal CIFAR-10 & 92.65* & 70.20 & 80.44\\
LDAM CIFAR-10 & 92.65* & 77.80 & 80.30 \\
C-ent SVHN & 94.91 & 83.29 & 85.60 \\
C-ent CelebA & 96.90 & 87.65 & 92.70\\

\bottomrule

\end{tabular}
\begin{tablenotes}
      \small
      \item {\footnotesize *Cross-entropy loss BAC.}
    \end{tablenotes}
\end{threeparttable}
\vspace{-.2cm}
%\end{table}
%\end{minipage}
%\end{wrapfigure}
\end{wraptable} 

Thus, a classifier re-trained with the \textit{latent embeddings} of the full, balanced training set is not able to recover the BAC of a combined CNN extractor trained on the same data.  This implies that the CNN extractor trained on imbalanced data has not learned the same latent features as the CNN extractor/classifier trained on balanced data. In the following experiments, we attempt to understand why this is the case.

\subsection{RQ2: what is the effect of imbalance on generalization?}
\label{sec:rq2_res}

\begin{figure}[h!]
   \vspace{-0.5cm}
  %\centering
  \subfloat[Balanced]{\includegraphics[width=0.5\textwidth]{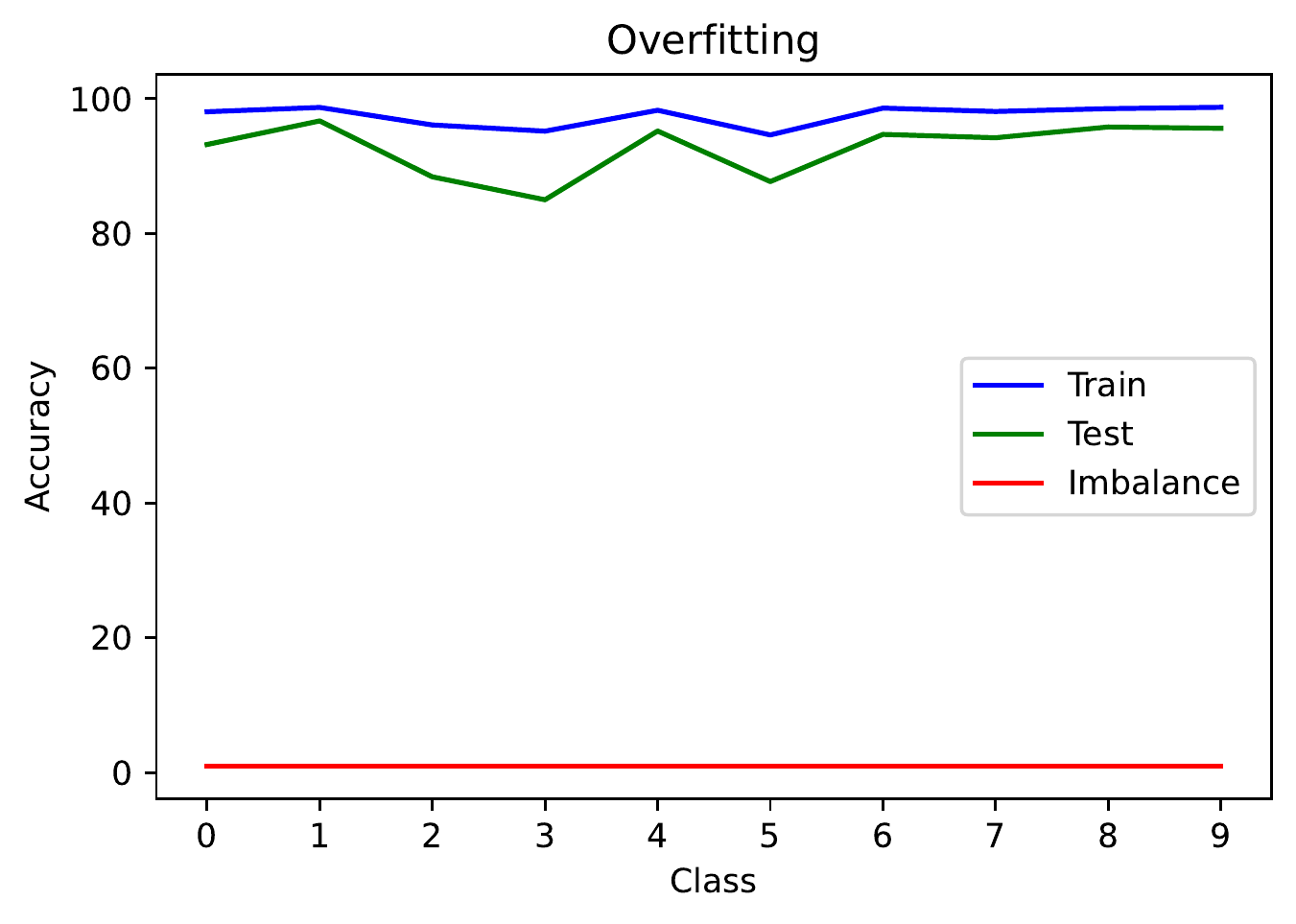}\label{fig:f1}}
  %\hfill
  \subfloat[Imbalanced]{\includegraphics[width=0.5\textwidth]{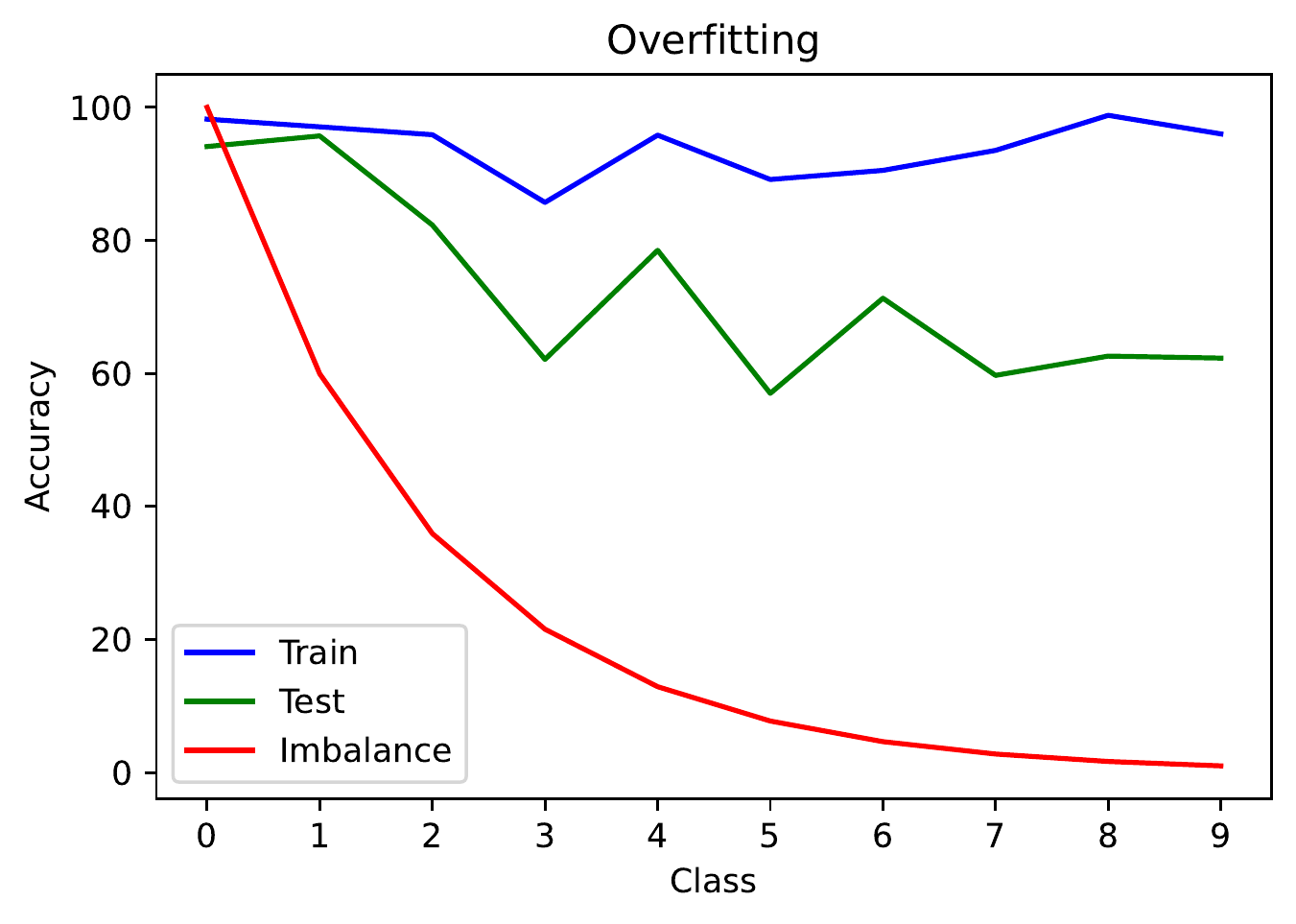}\label{fig:f2}}
  \caption{The figure on the left shows that a CNN can readily generalize from training to test distributions when trained with balanced CIFAR-10 data.  When the same model architecture is trained on imbalanced CIFAR-10 data, minority classes display much greater difficulty generalizing compared to majority classes.  In both diagrams, the red line indicates "inverse" class imbalance levels, such that, in the right diagram, class 9 is imbalanced 100:1 compared to class 0.}
  \label{fig_ofit}
  \vspace{-0.2cm}
\end{figure}

Here, we investigate a CNN's ability to generalize on balanced and imbalanced CIFAR-10 data.
Figure ~\ref{fig_ofit} shows that a CNN trained with a balanced CIFAR-10 training set is able to generalize from the training to the test distribution with relative ease. However, when the same dataset is imbalanced, the model displays both declining accuracy and increasing over-fitting for minority classes. In Figure ~\ref{fig_ofit}, the blue and green lines show training and test accuracy, respectively. The red line indicates the level of class imbalance for the CIFAR-10 dataset, which increases exponentially in the diagram on the right and is flat in the left diagram. For imbalanced data, the model is able to almost perfectly memorize the training data, but it has difficulty generalizing to the minority class test distribution. The same trend is repeated for the other datasets and cost-sensitive algorithms. 

\begin{wraptable}{L}{0.7\textwidth}
%\vspace{-1cm}
%\begin{table}[h!] %*
\vspace{-0.5cm}
%\small
%\scriptsize
\footnotesize
\caption{Effect of Imbalance on Generalization (BAC)}
\label{tab: genl}
\centering
\begin{tabular}{ p{1.7cm}p{1.2cm}
p{1.2cm}p{1.2cm}p{1.2cm}}
\toprule

\textbf{Description} & \textbf{Train Majority Class (Bal.)} & \textbf{Test Majority Class (Bal.)} & \textbf{Train Minority Class (Imbal.)} & \textbf{Test Minority Class (Imbal.)} \\

\midrule

C-ent CIF10 & 98.22 & 94.10 & 96.00 & 62.30 \\
Focal CIF10 & 99.48 & 96.00 & 98.00 & 39.30\\
LDAM CIF10 & 98.42 & 94.10 & 100.0 & 72.90 \\
C-ent SVHN & 99.78 & 98.13 & 100.0 & 61.47 \\
C-ent CelebA & 99.88 & 99.10 & 94.40 & 76.20\\
%\midrule

\bottomrule

\end{tabular}
\vspace{-.3cm}
%\end{table}
\end{wraptable}

Table ~\ref{tab: genl} shows BAC for the majority and minority classes, where the majority is the class with largest number of training examples and the minority is the class with the fewest. In all cases, the models have almost perfect accuracy on the training data, but have difficulty generalizing to the minority class test data, when the training sets are imbalanced.

\subsection{RQ3: does a CNN rely on top-K features?}
\label{sec:rq3_res}

%\begin{wrapfigure}{L}{0.5\textwidth}
\begin{figure}[h!]
   \vspace{-0.5cm}
  %\centering
  \subfloat[Balanced Coverage Ratio]{\includegraphics[width=0.5\textwidth]{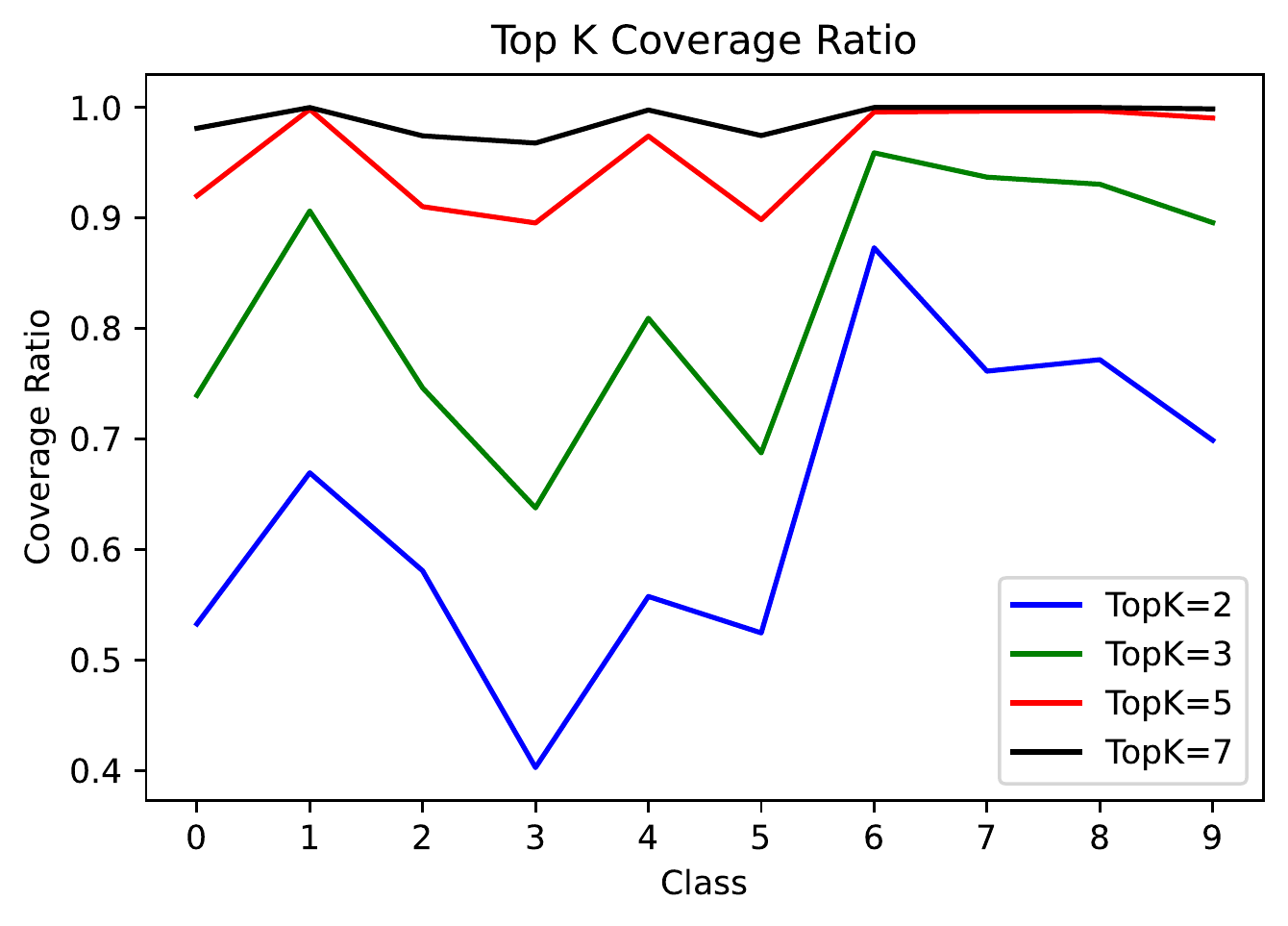}\label{fig:f3}}
  %\hfill
  \subfloat[Imbalanced Coverage Ratio]{\includegraphics[width=0.5\textwidth]{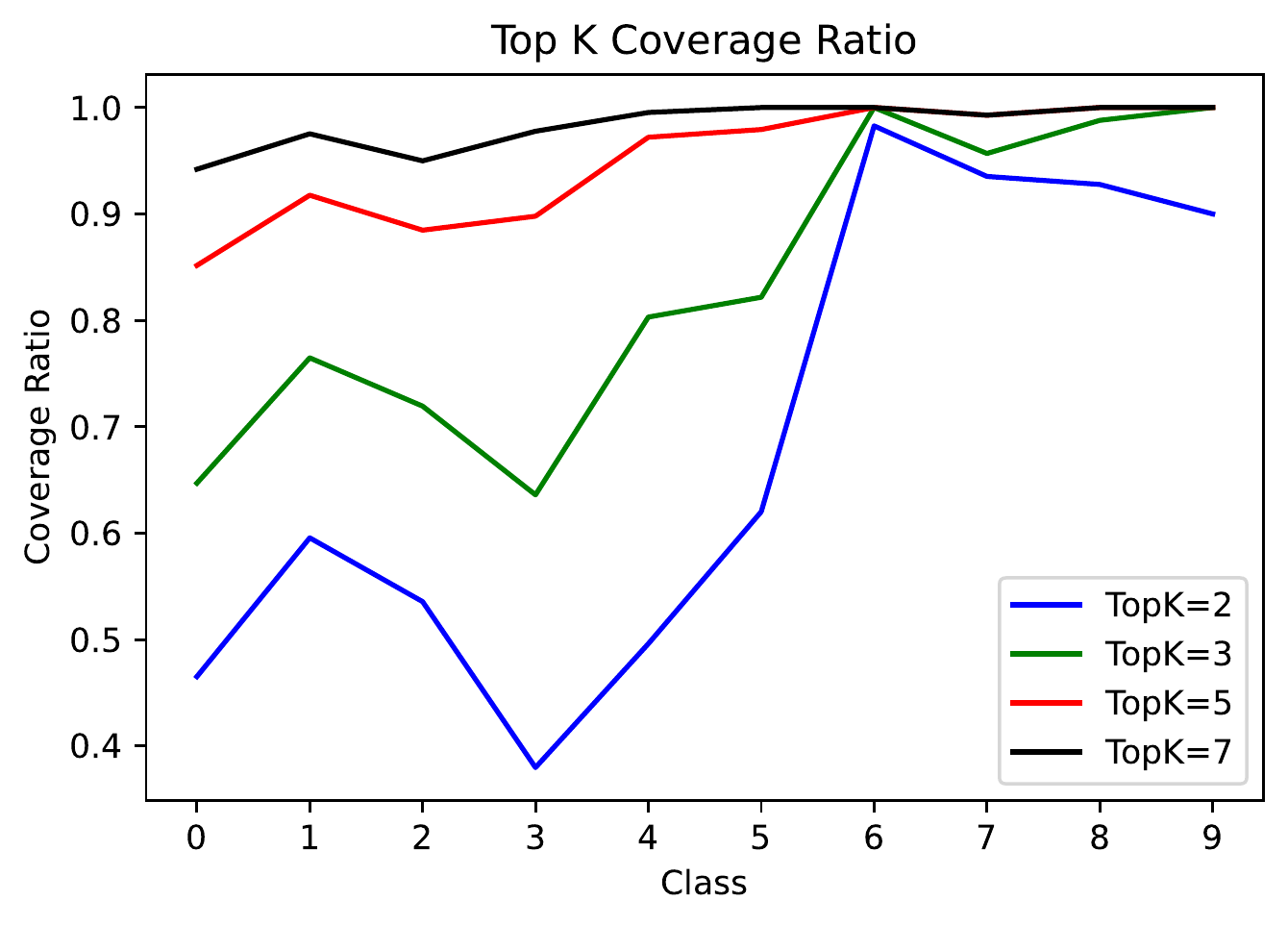}\label{fig:f4}}
  \caption{The figure on the left displays the \textit{class top-K coverage ratio} for a Resnet-32 with balanced CIFAR-10 data; and the one on the right shows imbalanced data.  In both cases, top $k = 7$ accounts for over 94\% of training set predictions.}
  \label{fig_c_ratio}
  \vspace{-0.3cm}
\end{figure}
%\end{wrapfigure}

Figure ~\ref{fig_c_ratio} shows the \textit{top-K coverage ratios} for two models: one trained with balanced, and the other trained with imbalanced, CIFAR-10 data for $K \in \{2,3,5,7\}$.  For the balanced data, there is no clear pattern for $K=2$ or $K=3$, denoted with blue and green lines, respectively.  For these K values, the top-K coverage ratio fluctuates between a low of 40\% and a high of 96\%, with no clear trend for the dataset as a whole.  For $K=3$, the model is better able to predict classes 6 to 9 compared to 0 to 5 on a balanced dataset; although the overall training set average remains low. However, at $K=5$ (red line), the ratio stabilizes between 89\% and 99\%; and at $K=7$ (black line), the model is able to predict a class label over 97\% of the time for all classes and training set examples on a balanced dataset.

Figure ~\ref{fig_c_ratio} reveals a different picture for imbalanced data.  There is a clearer trend based on the imbalance level.  For $K=2$, the model struggles to predict the majority classes (0 to 3) with only 2 features, 60\% of the time; however, there is a clearly sloping upward trend after that, with the model able to predict the 4 most extreme minority classes (6 to 9), with only 2 features over 90\% of the time. In contrast, on balanced data, the model performed well for class 6, but there is a downward sloping line for classes 7 to 9.  Similar to the balanced data, at $K=7$, the model is able to predict a class over 94\% of the time with only 7 features for all classes. Interestingly, $K=7$ is approx. 10\% of the total number of \textit{FE} and \textit{CE} per instance (i.e., 7 out of 64) for the Resnet-32 architecture. Since for both balanced and imbalanced datasets, $K=7$ constitutes the number of relevant features needed to make a prediction in over 94\% of the training instances, we focus on 7 features in subsequent experiments. 

A similar trend is reflected in other cost-sensitive algorithms and datasets.  In the case of LDAM and the focal loss, $K=2$ and $K=11$ constitute the number of relevant features necessary to predict 100\% and over 90\% of the training instances, respectively.  For CelebA and SVHN, $K=2$ and $K=3$ are needed to predict 100\% and over 94\% of training instances, respectively.  In all cases, K is far smaller than the dimension of the latent space (\textit{FE and CE}).

These results confirm that a CNN classifier relies on a limited number of features to make its prediction and this number is less than the dimension of the classification layer, such that $K<\!\!<\!H$.  These results also show that a CNN generally uses fewer features (\textit{CE}) to distinguish minority classes than majority classes.

\subsection{RQ4: does imbalance affect the diversity of learned features?}
\label{sec:rq4_res}

\begin{wrapfigure}{L}{0.5\textwidth}
%\begin{minipage}{0.42\textwidth}
%\begin{figure}[h!]
\vspace{-.3cm}
%[height=4.5cm,
\centering
%width  1.0  scale=0.5,
  \includegraphics[width=0.5\textwidth]{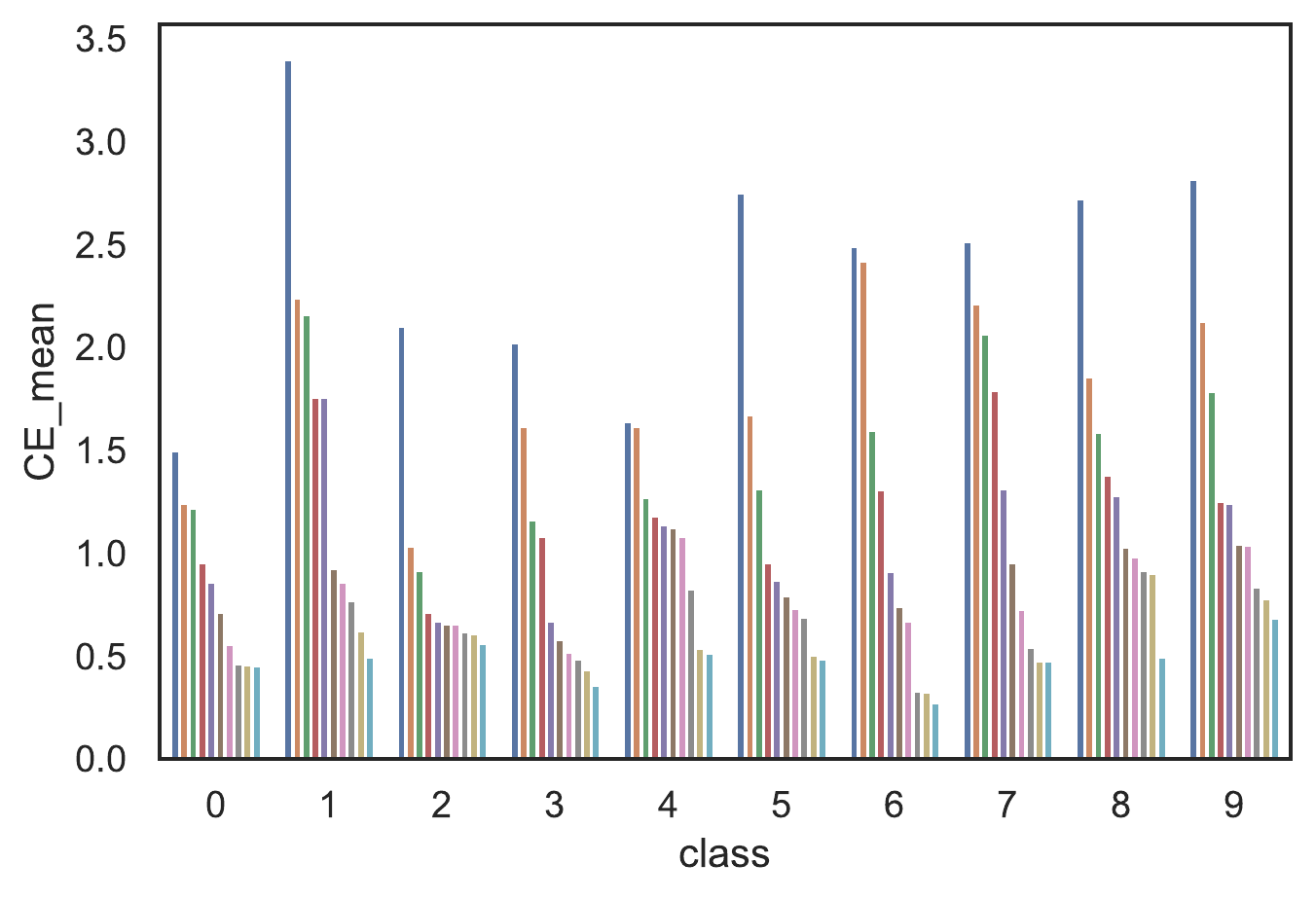}
  \caption{This figure show the 10 largest mean magnitudes of \textit{CE} for CIFAR-10 classes extracted from a CNN trained on balanced data. The \textit{CE} are sorted, with the \textit{CE} identity varying on the x-axis by class.  The shape of the histograms and the magnitude of the mean value ranges appear relatively similar for all classes.}
  \label{fig_c10_ce_mu_mag}
  \vspace{-0.5cm}
%\end{figure}
%\end{minipage}
\end{wrapfigure}

To gain a better understanding of why fewer features are required to distinguish  minority classes, we visualize the mean magnitudes of the top-K $ce$ for all classes. Figure ~\ref{fig_c10_ce_mu_mag} shows the ten largest mean magnitudes of $ce$ by class for a CNN trained on balanced CIFAR-10 data.  The mean magnitudes are sorted by class so that we can clearly see the range and scale of the values for the most significant features. For balanced data, the $ce$ for all classes reside in a narrow band between 0 and 3.4.  The single largest mean $ce$ in each class spans from 1.5 to 3.4.

\begin{wrapfigure}{L}{0.5\textwidth}
%\begin{figure}[h!]
\vspace{-.5cm}
%[height=10cm,
\centering
  \includegraphics[width=0.5\textwidth]{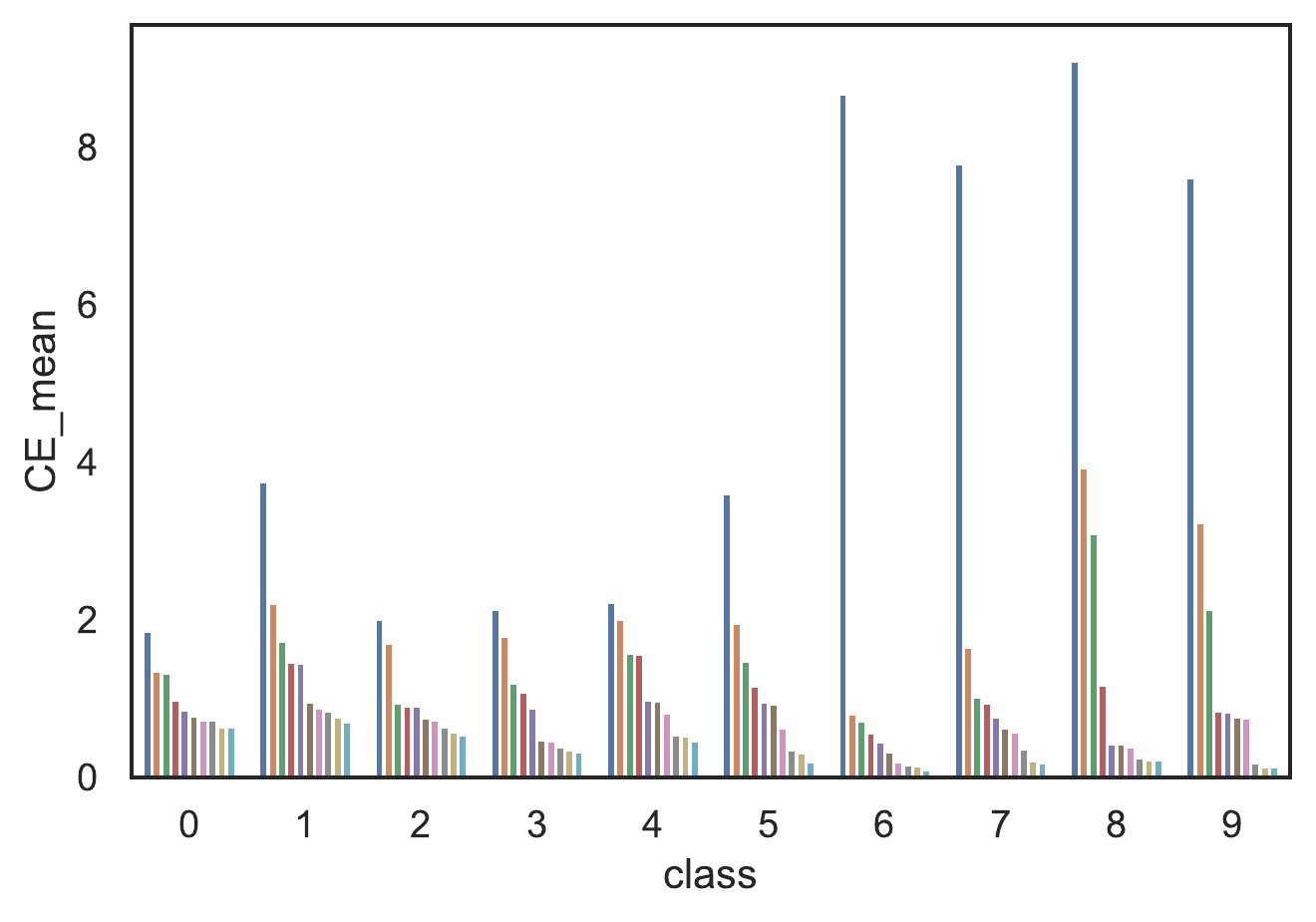}
  \caption{This figure shows the mean magnitudes of $ce$ for CIFAR-10 classes for a CNN trained on imbalanced data. The $ce$ are sorted, with the $ce$ identity varying on the x-axis by class. The extreme minority classes (6 to 9) exhibit a more narrow band of high valued mean features with higher overall magnitude.}
  \label{fig:ce_mu_mag_imb}
  \vspace{-0.5cm}
%\end{figure}
\end{wrapfigure}

In contrast, Figure ~\ref{fig:ce_mu_mag_imb} reveals a wide band between the mean magnitudes of the class $ce$ with the ten largest mean magnitudes of 0 to 9.1. For the imbalanced training set, the largest class $ce$ mean magnitude spans from 1.8 to 9.1, which is approx. triple the balanced data range. 

The $ce$ show a clear trend of large mean magnitudes for classes with few training examples and much smaller mean magnitudes for classes with many examples.  The single largest $ce$ for the extreme minority classes (6 to 9), with more than 20:1 imbalance, average 8.3, whereas the classes with more examples (0 to 5) average only 2.6.

The pattern of larger top-K CE mean magnitudes where $K=1$ is present in other datasets and cost-sensitive algorithms.  Table ~\ref{tab: ce_diffs} shows the mean magnitude of the largest single \textit{ce} for the majority class and the average for all other classes.  In all cases, the majority class \textit{ce} magnitude is at least 2X smaller than the minority classes. 

\begin{wraptable}{L}{0.63\textwidth}
%\vspace{-1cm}
%\begin{table}[h!] %*
\vspace{-0.5cm}
%\small
%\scriptsize
\footnotesize
\caption{Mean Magnitude Largest Class ce}
\label{tab: ce_diffs}
\centering
\begin{tabular}{ p{2.0cm}p{.9cm}
p{.9cm}p{.9cm}p{.9cm}}
\toprule

\textbf{Description} & \textbf{CelebA} & \textbf{SVHN} & \textbf{LDAM} & \textbf{Focal} \\

\midrule

Majority Cls. & 0.5897 & 1.4899 & 4.3578 & 0.8553 \\
Avg. Other Cls. & 1.4945 & 3.4955 & 12.5248 & 3.4335\\
Ratio & 2.53 & 2.35 & 2.87 & 4.01 \\
%\midrule

\bottomrule

\end{tabular}
\vspace{-.5cm}
%\end{table}
\end{wraptable}

%\begin{figure}[h!]
%\vspace{-0.2cm}

%\centering
%  \includegraphics[width=0.45\textwid%th]{cif10_diffs_ce_mu.pdf}
%  \caption{This figure shows the norm difference in the mean magnitudes of the class top 10 $ce$ between balanced and imbalanced CIFAR-10 data.  There is a clear trend of larger differences as imbalance increases.}
%  \label{fig:ce_mu_diff}
%  %\vspace{0.5cm}
%\end{figure}

There is also a greater and faster drop off in the mean magnitudes of $ce$ for minority classes, after the single largest class $ce$.  As class imbalance increases, the mean magnitudes of the single largest $ce$ increase and there is greater concentration of large responses in only a handful of $ce$.  This drop off is clearly shown in 
Figure ~\ref{fig:ce_mu_mag_imb} for CIFAR-10. As imbalance grows, fewer features with higher mean magnitudes contribute to the classifier's prediction.

\begin{figure}%[h!]
\vspace{-.5cm}
%[height=10cm,
\centering
  \subfloat[Balanced Coverage Ratio]{\includegraphics[width=0.5\textwidth]{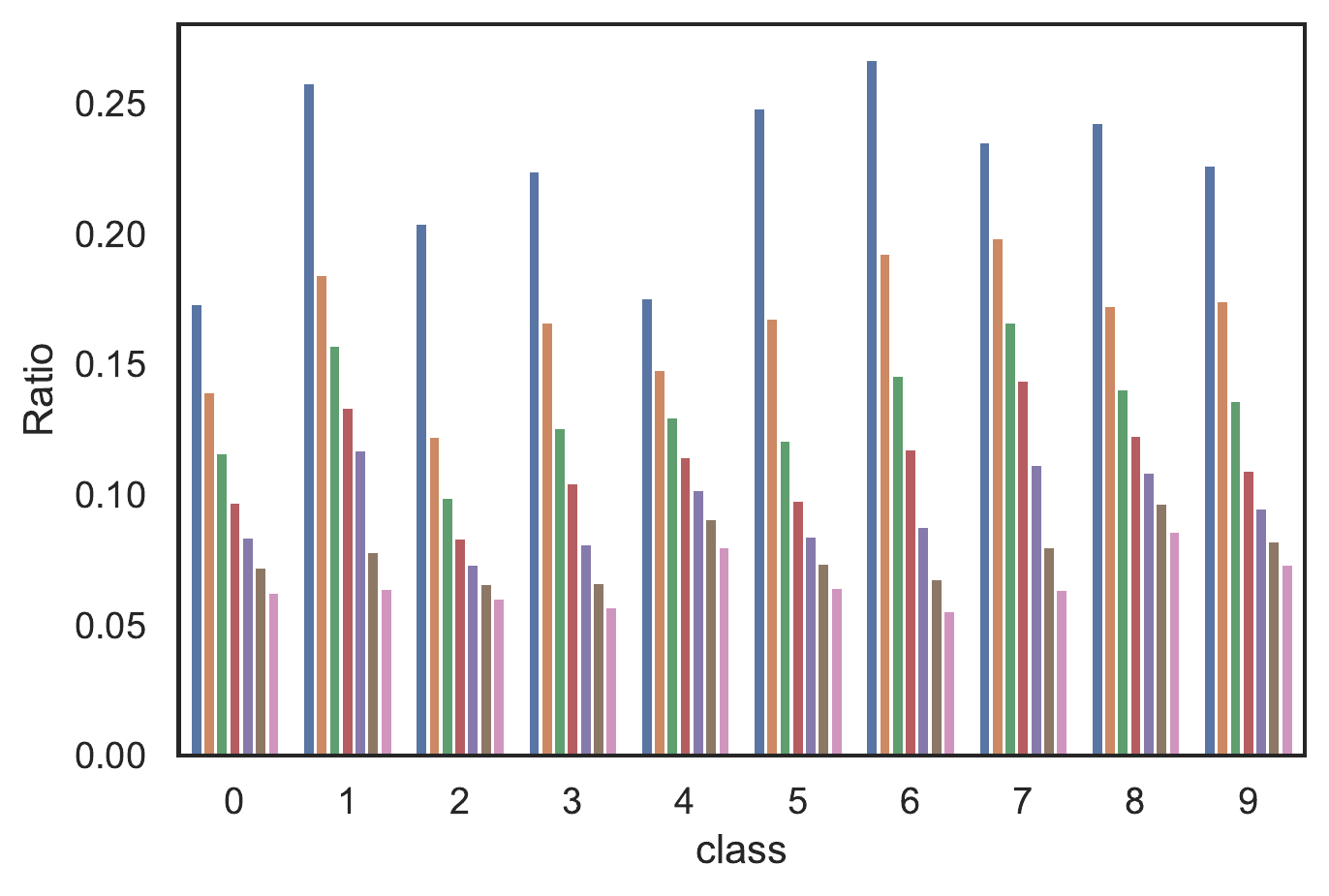}\label{fig:f5}}
  %\hfill
  \subfloat[Imbalanced Coverage Ratio]{\includegraphics[width=0.5\textwidth]{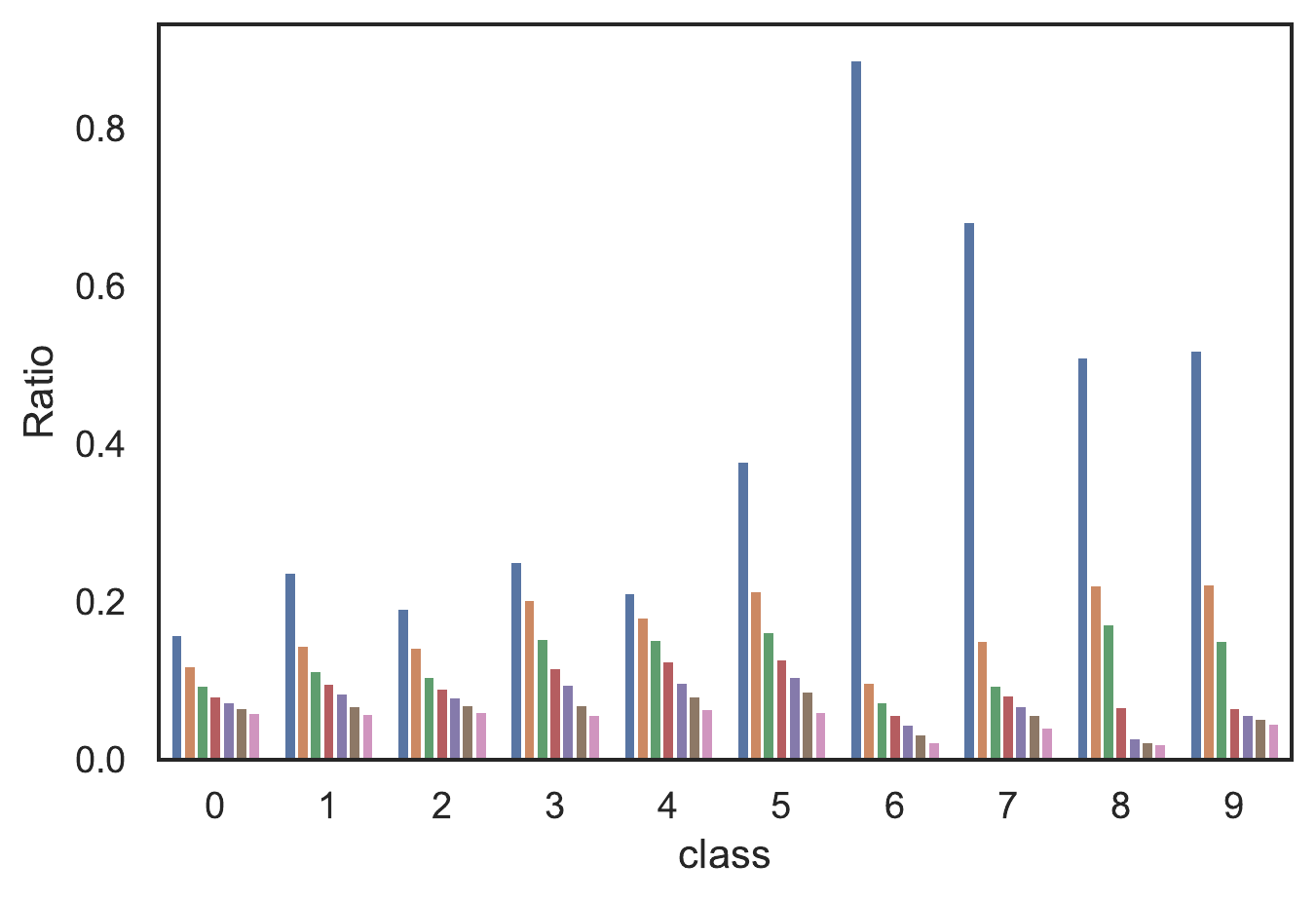}\label{fig:f6}}
  \caption{This figure shows the percentage that the top 7 $ce$, by class, contribute to each logit instance for a balanced and imbalanced CIFAR-10 dataset.  The percentage is based on averaging all class instances.  For the imbalanced data, fewer $ce$ contribute a greater percentage to the prediction logit.}
  \label{fig:ce_bal_imb_logits}
  \vspace{-0.5cm}
\end{figure}

Figure ~\ref{fig:ce_bal_imb_logits} shows the percentage that the top 7 $ce$, by class, contribute to each logit instance for a balanced and imbalanced CIFAR-10 dataset.  Each $ce$ percentage is based on averaging all class instances.  For the imbalanced data, fewer $ce$ contribute to a greater percentage of the prediction logit.  For balanced data, in the left diagram,  \textit{no single} $ce$ contributes more than 26\% of the predicted logit.  However, in the right diagram, which depicts imbalanced data, there are \textit{5} classes that have $ce$ that contribute more than 35\% to the predicted logit.
This trend is repeated for other datasets and cost-sensitive algorithms.  Table ~\ref{tab: ce_logit_contrib} shows the contribution of the single largest ce to the class logit for the majority class and all other classes.  In the case of the majority class, it's largest logit contributes between 2-4 times less to the overall class logit, which indicates that the majority class relies on a wider diversity of features to arrive at its class decision.

\begin{wraptable}{L}{0.63\textwidth}
\vspace{-.5cm}
%\begin{table}[h!] %*
%\vspace{-0.8cm}
%\small
%\scriptsize
\footnotesize
\caption{Top \textit{ce} Contribution to Class Logit}
\label{tab: ce_logit_contrib}
\centering
\begin{tabular}{ p{2.0cm}p{.8cm}
p{.8cm}p{.8cm}p{.8cm}}
\toprule

\textbf{Description} & \textbf{CelebA} & \textbf{SVHN} & \textbf{LDAM} & \textbf{Focal} \\

\midrule
Majority Cls. & .1191 & .1191 & 6.121 & .1267 \\
Avg. Other Cls. & .5220 & .2774 & 12.01 & .3844\\
Ratio & 4.38 & 2.33 & 1.96 & 3.03 \\

%\midrule

\bottomrule

\end{tabular}
\vspace{-.5cm}
%\end{table}
\end{wraptable}

%\begin{figure}[h!]
%%\vspace{-1.2cm}
%%[height=10cm,
%\centering
%  \includegraphics[width=0.45\textwidth]{cif10_bal_imb_ce_cnt_logits_.pdf}
%  \caption{This figure counts the number of $ce$ by class required to make make up 70\% or greater of the predicted logit.  Fewer are required as class imbalance grows, revealing a narrower class decision.}
%  \label{fig:ce_logit_count}
%  %\vspace{0.5cm}
%\end{figure}

Collectively, these results indicate that a CNN classifier forms its decisions on a small portion of the dimension of its feature inputs.  In the case of minority classes, the number of relevant features is even smaller (2 or 3 in some cases). We conjecture that the number of relevant features is wider for majority classes because their examples are more diverse (i.e., there are a larger number of relevant features per class that each individually contribute smaller size magnitudes to the logit).  Because the majority distribution is more diverse, the model requires a larger decision manifold (more relevant features) to distinguish the class instances, which cumulatively add up to the logit.  In contrast, due to modest minority class diversity, the model generates only a few, high valued response $ce$ to distinguish these classes.

In the next two subsections, we will consider whether the model weights $W_C$ or the learned feature embeddings \textit{FE} are responsible for the narrow, high-valued $ce$ responses of minority classes. 

\subsection{RQ5: how significant are classifier weights vs. features to the network's prediction?}
\label{sec:rq5_res}

\begin{wrapfigure}{L}{0.5\textwidth}
%\begin{figure}[h!] %*
   \vspace{-.5cm}
  %\centering
  \subfloat[Class Weight Magnitudes: Top 10 by Class]{\includegraphics[width=0.45\textwidth]{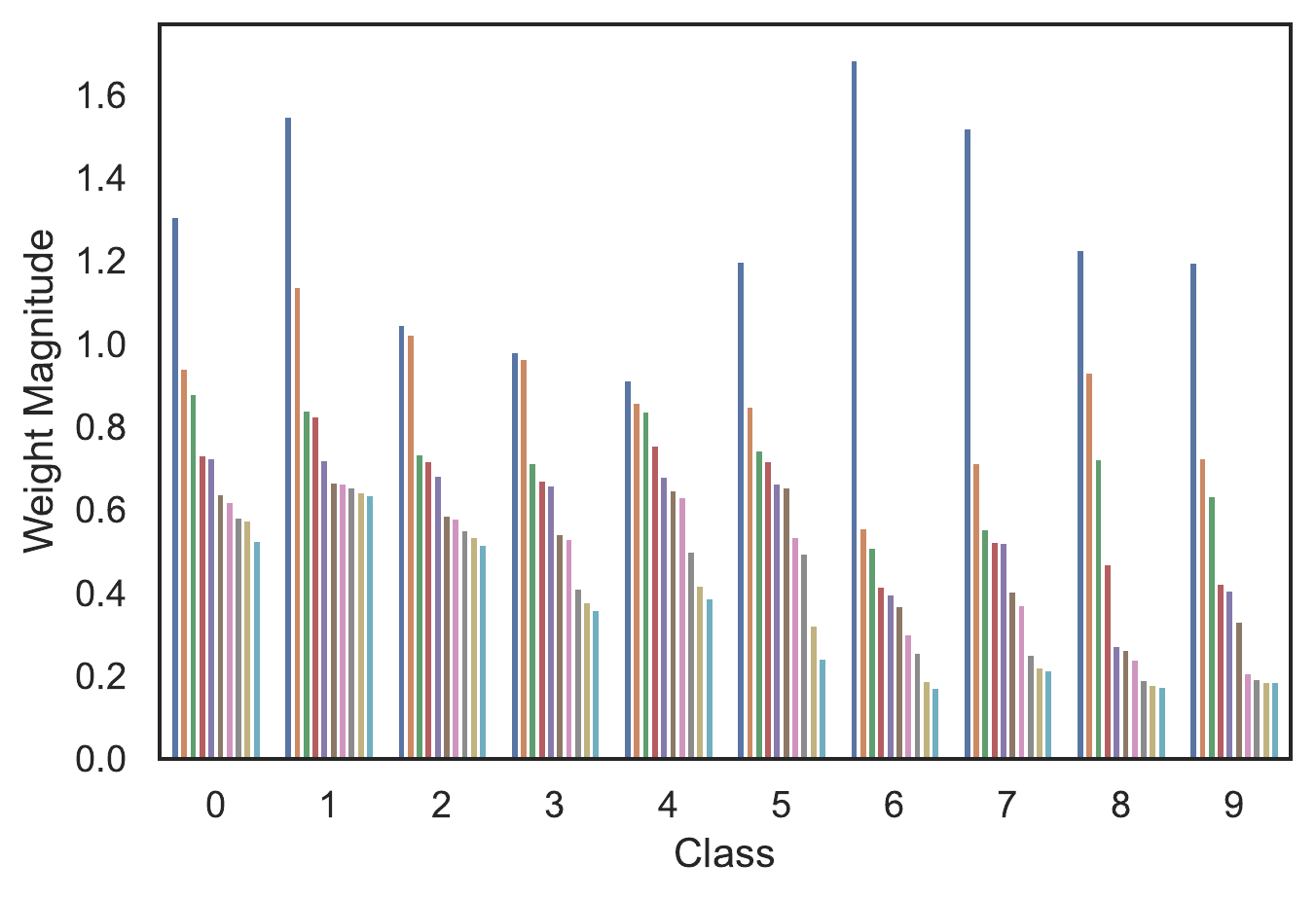}\label{fig:f7}}
  %\hfill
  %\subfloat[Class Sums for Top 10 Weights ]{\includegraphics[width=0.45\textwidth]{cif10_imb_wt10_sums.pdf}\label{fig:f8}}
  \caption{This figure shows the magnitudes of the ten largest weights, $W_c$, by class, for imbalanced CIFAR-10 data. The majority classes have a wider cross-section of larger weights, whereas the minority class large magnitudes are concentrated in fewer weights.}
  %The figure on the right shows the sum of the top 10 weights by class.  The majority class weight sums are larger than the minority classes, indicating that the majority class prediction is weighted on a larger cross-section of features.}
  \label{fig_imb_wts}
  \vspace{-0.5cm}
%\end{figure}
\end{wrapfigure}

Figure ~\ref{fig_imb_wts} shows the ten largest weight mean magnitudes, $W_c$, by class, for imbalanced CIFAR-10 data. The majority classes have a wider cross-section of larger weights, whereas the minority class has a narrower concentration.  The larger majority class weight mean magnitudes can be seen by comparing the sum of the class top 10 weight mean magnitudes for the majority and minority classes.  

For the CIFAR-10 dataset trained with cross-entropy, focal loss and LDAM, and for SVHN and CelebA, the sum of the majority class top 10 $W_C$ mean magnitudes are 7.54, 4.89, 11.68, 7.12, and 5.51, respectively.  In contrast, for the minority class, the sum of the top 10 $W_C$ mean magnitudes are 4.50, 3.48, 5.80, 3.94, and 4.38, respectively. 

The weight sums are significant because the classifier arrives at its decision by summing the element-wise multiplication of weights and \textit{FE}. We conjecture that there is a wider cross-section of larger weights in majority classes because the class top-K $fe$ are more diverse than in the case of the majority.  The model has learned more diverse features for the majority due to more varied examples and it must weight these more frequently occurring features to distinguish majority instances. 

Although the weights are clearly biased toward the majority, the magnitude of the weights does not account for the large magnitudes of the class top 10 $ce$ members.  For example, in the case of the extreme minority classes (8 and 9) for CIFAR-10, their top $ce$ have mean magnitudes greater than 8.0, yet the corresponding weights are only approx. 1.2 (see Figures ~\ref{fig_c10_ce_mu_mag} and ~\ref{fig_imb_wts}). A similar trend is evident in other datasets and cost-sensitive algorithms. The largest mean $W_C$ is 1.29, 2.36, 1.25, and 1.04 for focal loss, LDAM, SVHN, and CelebA, respectively.  However, the largest mean \textit{ce} are 8.77, 13.32, 6.50, and 1.49 for the same datasets. 

This implies that weight re-balancing strategies employed by some cost-sensitive, over-sampling, or classifier re-training methods may not be sufficient to redress the class imbalance problem.  Although weight re-balancing may be helpful, there may be limits to the amount of class bias that it can address due to the scale difference between the weights ($W_C$) and \textit{CE} values. 

Because $W_C$ appear to only have a minor impact on minority class \textit{CE}, we next examine its other component, \textit{FE}.

\subsection{RQ6: are majority class features more diverse?}
\label{sec:rq6_res}

In this section, we take a closer look at \textit{FE}, which is the other component of \textit{CE}, and investigate why there is a greater concentration of high valued feature responses for minority class \textit{CE} compared to majority class \textit{CE}.  

\begin{wrapfigure}{L}{0.5\textwidth}
%\begin{figure}[h!]
%\vspace{-1.2cm}
%[height=4.5cm,
\centering
  \includegraphics[width=0.47\textwidth]{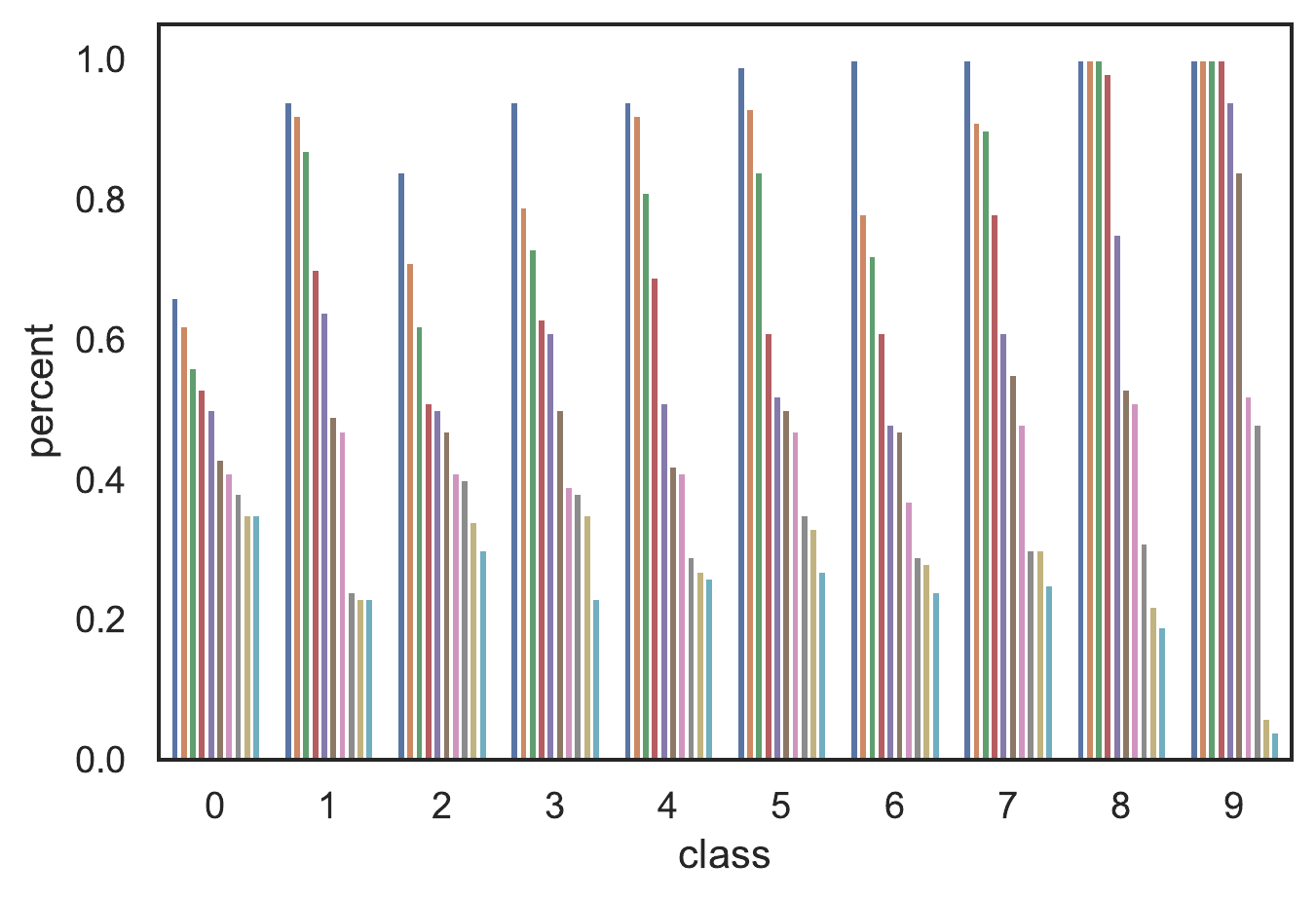}
  \caption{This figure shows the \textit{class top-K FE} ratio for imbalanced CIFAR-10 data.  It conveys the diversity of the most frequently occurring \textit{top-K fe} in each class.  The extreme majority class (0) shows no top-K class ratios greater than 67\%, whereas the extreme minority classes (8 \& 9) have 4 and 5 features ($fe$), respectively, that are present in over 90\% of class instances,}
  \label{fig:fe_diversity}
  \vspace{-0.1cm}
%\end{figure}
\end{wrapfigure}

Figure ~\ref{fig:fe_diversity} shows the \textit{class top-K fe coverage ratios} for the ten most frequently occurring features (fe) per class.  The \textit{FE} were extracted from a CNN trained on imbalanced CIFAR-10 data.  Low values indicate that a larger number of varied features are needed to distinguish a class.  In the figure, the extreme majority class (0) shows no class top-K \textit{fe} coverage ratio greater than 67\%, whereas the extreme minority classes (8 \& 9) have 4 and 5 \textit{fe}, respectively, that are present in over 90\% of class instances. 

\begin{wrapfigure}{L}{0.5\textwidth}
%\begin{figure}[h!]
%\vspace{-1.2cm}
%[height=4.5cm,
\centering
  \includegraphics[width=0.47\textwidth]{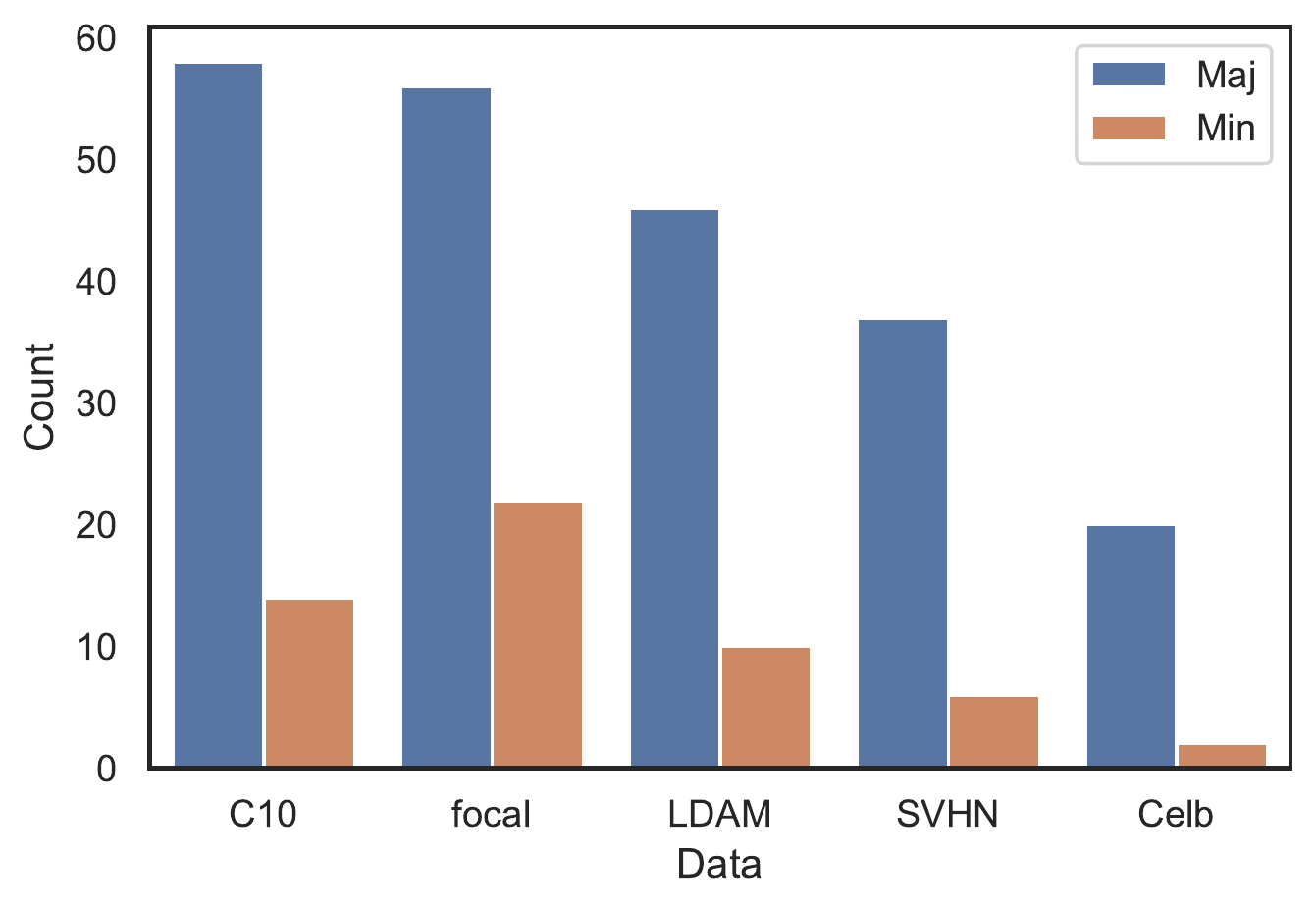}
  \caption{This figure shows the number of \textit{class top K FE} that are necessary to describe all instances in the majority and minority classes for the CIFAR-10  dataset with the cross-entropy, focal and LDAM loss functions, and the SVHN and CelebA datasets using cross-entropy loss.  In all cases, it requires significantly more \textit{class top-K FE} to describe the majority class than the minority class.}
  \label{fig:fe_counts}
  \vspace{-0.1cm}
%\end{figure}
\end{wrapfigure}

Figure ~\ref{fig:fe_counts} shows the number of top-K class \textit{fe} that are required to fully describe all class instances for majority and minority classes. For cost-sensitive algorithms and all three datasets, fewer top-K are required for minority classes.

Together, Figures ~\ref{fig:fe_diversity} and ~\ref{fig:fe_counts} demonstrate that it takes fewer features to describe minority than majority classes. Due to the greater diversity of majority instances, a greater number of features are needed to predict the full class. 

Figure ~\ref{fig_feat_mu} (a) shows the ten largest \textit{fe} mean magnitudes, by class, for an imbalanced CIFAR-10 training set.  The scale of these magnitudes more closely aligns with the ten largest mean magnitudes of imbalanced $ce$ shown in Figure ~\ref{fig_c10_ce_mu_mag} than the $W_C$ in Figure ~\ref{fig_imb_wts}, and demonstrates that \textit{FE} have a relatively larger impact on \textit{CE} (i.e., the model's decision) than $W_C$.  This observation implies that, in order to influence \textit{CE}, a method must modify the \textit{FE} extracted by a CNN and somehow augment the diversity of the initial, more static minority classes.  However, such a task is not easy, since the test distribution or its diversity cannot be known in advance.

\subsection{RQ7: are false positives an indicator of network memorization of training data?}
\label{sec:rq7_res}

For RQ7, we examine how the latent features (\textit{FE}) that a CNN has learned affects its ability to generalize to the minority class test distribution. We compare the model's internal embeddings (\textit{FE}) in the train, test true positive (TP) and test false positive (FP) sets so that we can identify differences in its internal embeddings when it makes correct versus incorrect predictions.

\begin{figure}[b!]
%\begin{figure*}[t!]
   %\vspace{-1.cm}
  %\centering
  \subfloat[Train: Means Top 10 fe]{\includegraphics[width=0.33\textwidth]{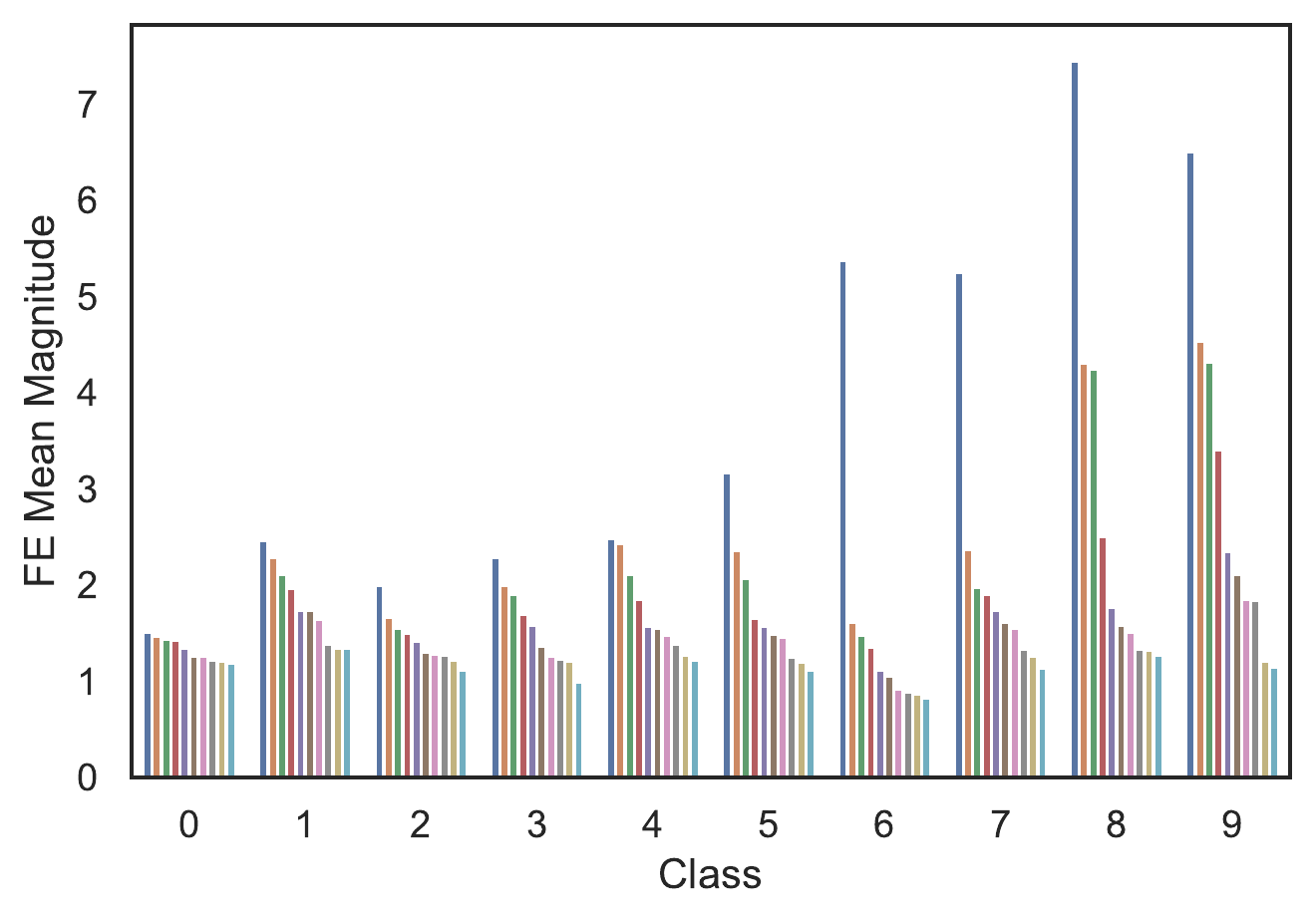}\label{fig:f9}}
  \hfill
  \subfloat[Test TP: Means Top 10 fe]{\includegraphics[width=0.33\textwidth]{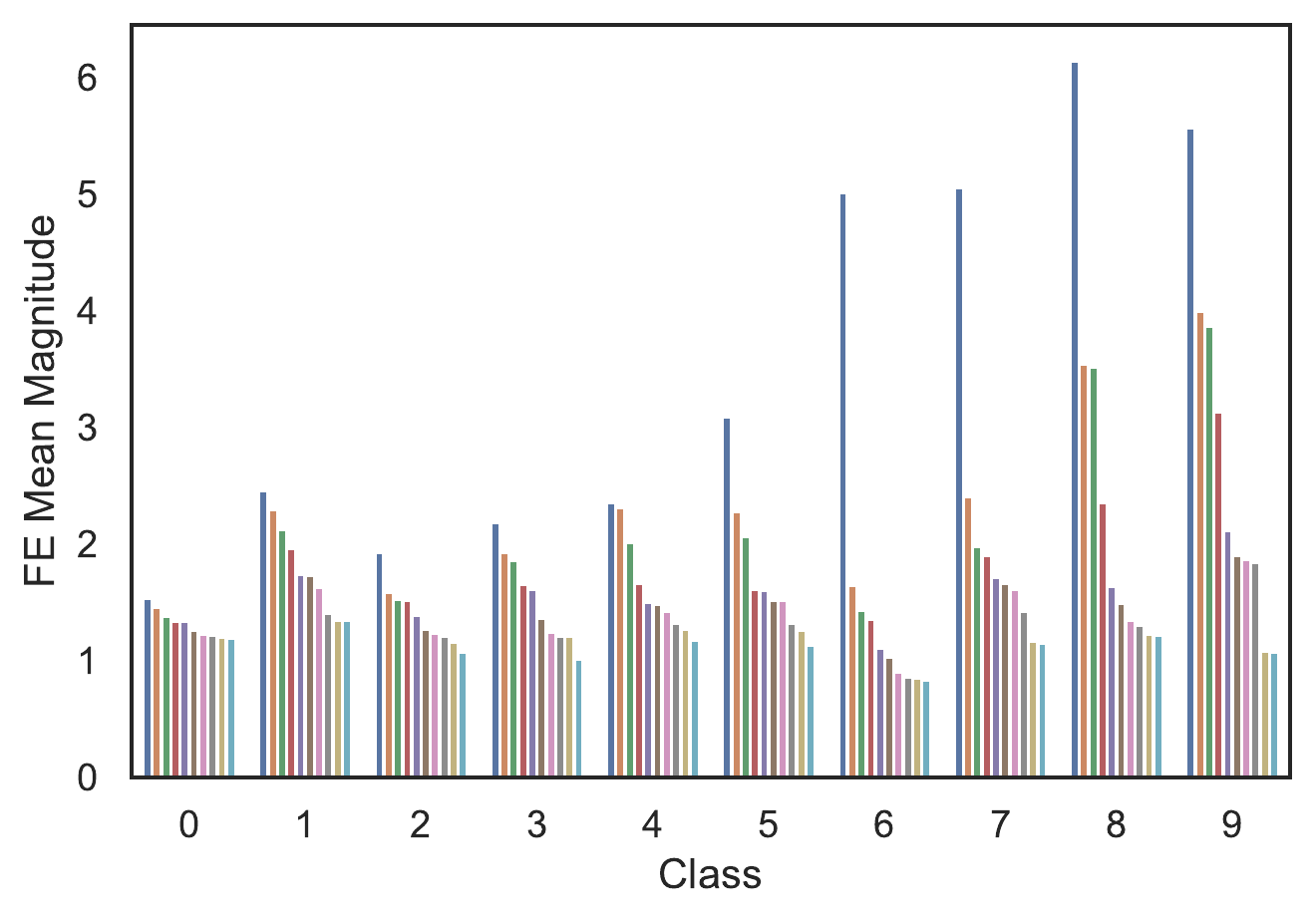}\label{fig:f10}}
  \hfill
  \subfloat[Test FP: Means Top 10 fe]{\includegraphics[width=0.33\textwidth]{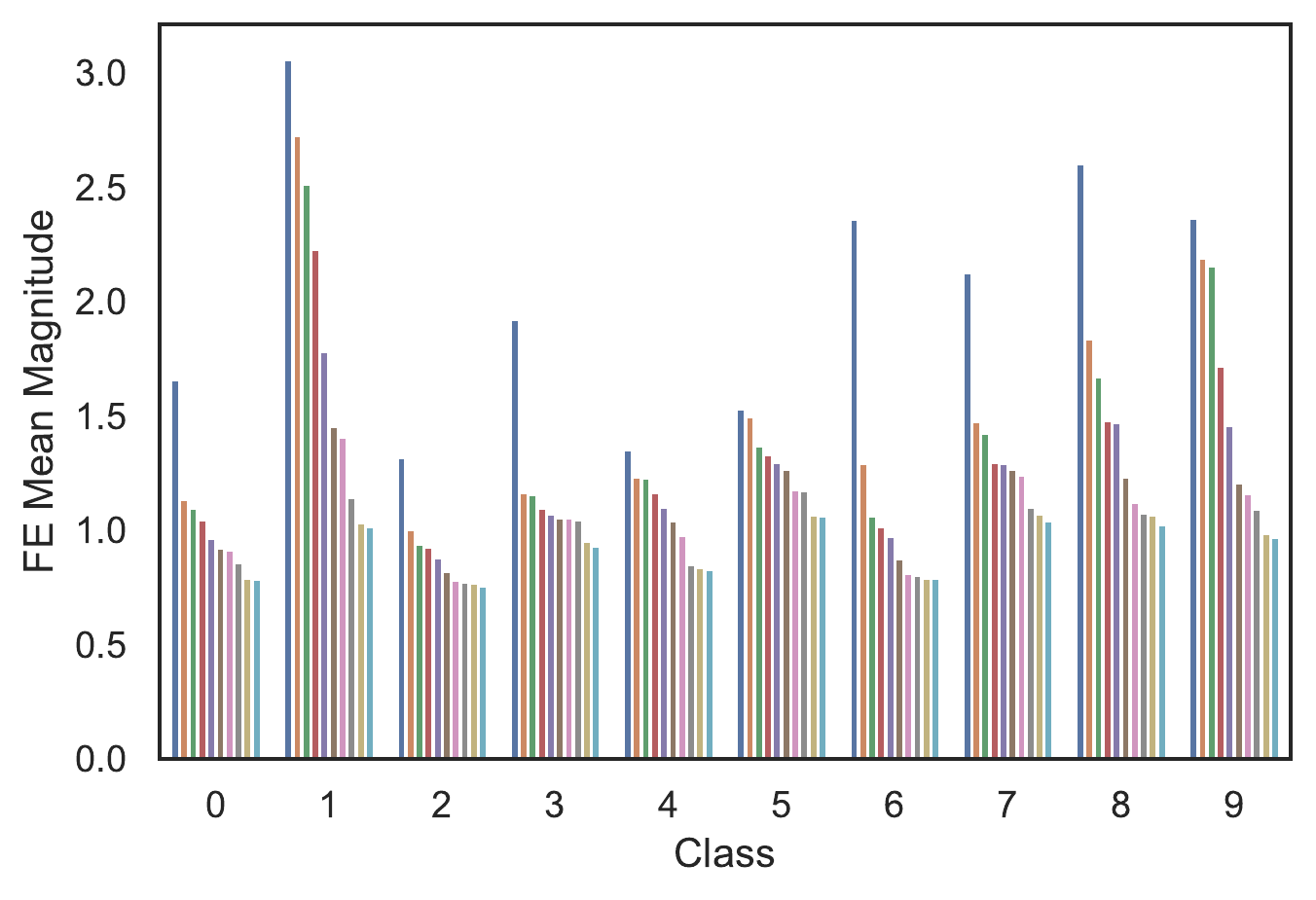}\label{fig:f11}}
  
  \caption{These figures show a clear divergence between the mean magnitudes of the class top 10 $fe$ members in the training and the test false positive set; however, many of the mean magnitudes for minority class top $fe$ are approx. half of their training set values.  In contrast, there is relatively close alignment between the training and test true positives}
  \label{fig_feat_mu}
  \vspace{-0.1cm}
\end{figure}
%\end{wrapfigure}

For CIFAR-10 trained on cross-entropy, in the case of true positives, there is a close correlation between the mean magnitudes of the features learned in training and the features in the test set.  For both \textit{CE} and \textit{FE}, there is 95\% and 96\% intersection between the top 10 most frequently occurring features in the train and test TP sets.  In the case of false positives, there is still relatively high correspondence between the identity of \textit{FE} or 70\% alignment; however, in the case of \textit{CE}, the correspondence drops to only 39\%.  In other words, whether the model makes correct or incorrect predictions, it basically relies on the same group of input features (FE) by class as it identified during training; however, there is a wide divergence between the final \textit{CE} used to make correct and incorrect predictions when compared to training. 

In order to gain insight into why this might occur, we look at the mean magnitude of the \textit{FE} in the training and test sets for a model trained on CIFAR-10 and cross-entropy.  Figure ~\ref{fig_feat_mu} shows a  relatively close alignment between the top $fe$ mean magnitudes of training and true positives.  In contrast, the same figure shows a clear \textit{divergence} between the mean magnitudes of the class top 10 $fe$ in the training set and the test false positive set, where many of the mean magnitudes for minority class top $fe$ are approx. half of their training set values. This visual observation is confirmed by the Frobenius norm (FB) of the mean magnitudes of $fe$ ($\mathit{FB}[\mu(Train_{FE}) - \mu(Test_{TP-FE})]$ and $\mathit{FB}[\mu(Train_{FE}) - \mu(Test_{FP-FE})]$). The Frobenius norm is 2.36 for training and test TPs and 12.42 for training and test FPs. The larger FB for training and test FPs show that the mean magnitude of the FPs are not well aligned with the training set, which affects the ability of the model to generalize.  

We repeated this exercise for the other datasets and cost-sensitive loss functions, with similar results.  The FB norm for training and test set TPs is 1.73, 4.01, 4.67, and 0.64 for focal loss, LDAM, SVHN, and CelebA, respectively.  In contrast, the FB norms are much higher for the training and test set FPs.  The FB norm for training and test FPs are 9.00, 21.16, 17.58, and 2.75 for the focal loss, LDAM, SVHN, and CelebA, respectively. 

The minority class true positive decision is based on a narrower group of class top K $fe$ that have high mean magnitudes and lower $W_C$ (hence, the logit is based on the sum of a few high magnitude $fe$ and weights).  We can imagine that if a model is biased to identify minority instances only when a narrow set of high valued features are present that it may harm its ability to generalize to minority class test examples that do not exhibit these characteristics. 

Collectively, these results show that the model is able to generalize from the training to test distributions when there is very close correspondence between the \textit{identity} of the most relevant features and the \textit{range} of their values (training and test TPs).  However, the model has difficulty generalizing when the range of $fe$ differs between the training and test sets (FPs), even when there is large (70\%) overlap in the identity of the class top K $fe$.

\section{Lessons learned}
\label{sec:les}

In this section, we summarize the key take-aways from our experiments.

First, CNNs trained with cross-entropy loss in a supervised manner are heavily reliant on carefully balanced training sets to achieve high accuracy.  This is consistent with, and confirms, other research \cite{bauder2018empirical,weiss2001effect,estabrooks2004multiple}.  Reducing the number of samples in the minority classes increases classifier bias toward the majority.  See RQ1 results in Section ~\ref{sec:rq1_res}.

Second, it is not clear that a CNN has learned the intrinsic features that define a class, but rather, high frequency patterns that occur in a sufficiently large number of training instances. Because the model has learned statistically frequent patterns in data, instead of intrinsic, compositional properties, it requires a diverse set of examples to find a sufficient number, and range, of latent feature magnitudes, to generalize from the training to the test set.  When a minority class is characterized by a low number of latent features in a lower response range in a test set, the model struggles to generalize to more diverse latent features. See Sections ~\ref{sec:rq6_res} and ~\ref{sec:rq7_res}.

Third, the magnitude or response that a CNN assigns to a feature has a large impact on CNN classification performance on imbalanced data.  CNNs trained on cross-entropy loss appear to assign high magnitudes to a narrow range of minority features and lower magnitudes to a larger number (more diverse) set of majority features.  This causes a disconnect during inference if the model is presented with minority class latent features (FE) that span a lower range during test than training, even if the features have the same identity. This observation confirms the brittleness of CNN latent embedding learning, which has been demonstrated in adversarial learning research \cite{ilyas2019adversarial,wang2020high}. See Section ~\ref{sec:rq4_res}.

Fourth, this paper postulates that the central problem of imbalanced image data lies in greater diversity for majority class latent features.  Imbalanced learning solutions that mainly target class number re-balancing, classifier retraining, increasing the cost of minority examples, or increasing the margin on class decision boundaries may plateau at some point. It is also not clear if merely over-sampling the minority class with interpolative, same-class examples is sufficient to re-dress lack of class diversity, as expressed by the number and magnitude of latent feature embeddings.  See Section ~\ref{sec:rq5_res}.

Finally, a CNN trained on cross-entropy or a cost-sensitive variant has difficulty generalizing if the magnitude of its top-K latent features in the training set do not match the test set. Effectively, a CNN memorizes training latent features in the form of model parameters, and if the response range of the features produced by these parameters and the input differs in the test set, then the model produces false positives. See Section ~\ref{sec:rq7_res}. 

\section{Conclusion}
\label{sec:concl}

CNNs are increasingly being deployed on real-world data, which is naturally skewed.  Training CNNs on imbalanced image data remains an open challenge. In this paper, we take steps toward demystifying a neural network's decision process for under-represented classes. By better understanding the role that a model's latent features play in its decision process, we aim to further research that improves a CNN's ability to generalize with respect to minority classes.

\backmatter

%\bmhead{Supplementary information}

%\bmhead{Acknowledgments}

\section*{Declarations}
 D. Dablain performed work on the paper during an internship with the Office of Naval Research.

\bibliography{sn-bibliography}% common bib file


\end{document}